%% file: main.tex
\documentclass{article}
\pdfoutput=1  

\PassOptionsToPackage{numbers, compress}{natbib}
\usepackage[preprint]{neurips_2024}  

\usepackage[utf8]{inputenc} 
\usepackage[T1]{fontenc}    
\usepackage{hyperref}       
\usepackage{url}            
\usepackage{booktabs}       
\usepackage{amsfonts}       
\usepackage{nicefrac}       
\usepackage{microtype}      
\usepackage{xcolor}         

\usepackage{bm}
\usepackage{amsmath,amssymb,mathtools}
\usepackage{graphicx} 
\usepackage{caption} 
\usepackage[subrefformat=parens]{subcaption}
\usepackage{comment}
\usepackage{listings}  

\RequirePackage{algorithm}
\RequirePackage{algorithmic}

\input{math_commands}
\input{symbols}

\title{Hierarchical Associative Memory, Parallelized MLP-Mixer, and Symmetry Breaking}

\author{%
    Ryo Karakida  \\
    AIST  \\
    \And
    Toshihiro Ota  \\
    CyberAgent  \\
    \And
    Masato Taki  \\
    Rikkyo University  \\
}

\begin{document}

\maketitle

\input{abstract}
\input{sec1}
\input{sec2}
\input{sec3}
\input{sec4}
\input{sec5}
\input{sec6}
\input{sec7}

\bibliography{references}
\bibliographystyle{nips}

\newpage
\appendix
\input{appendix}

\end{document}

%% file: math_commands.tex
\usepackage{amsmath,amsfonts,bm}

\def\Figref#1{Figure~\ref{#1}}

\def\eqref#1{equation~\ref{#1}}

\def\algref#1{algorithm~\ref{#1}}

\def\1{\bm{1}}

\def\eps{{\epsilon}}

\def\rb{{\textnormal{b}}}

\def\rh{{\textnormal{h}}}

\def\rt{{\textnormal{t}}}

\def\vb{{\bm{b}}}

\def\vh{{\bm{h}}}

\def\vt{{\bm{t}}}

\DeclareMathAlphabet{\mathsfit}{\encodingdefault}{\sfdefault}{m}{sl}
\SetMathAlphabet{\mathsfit}{bold}{\encodingdefault}{\sfdefault}{bx}{n}

\newcommand{\Ls}{\mathcal{L}}
\newcommand{\R}{\mathbb{R}}

\newcommand{\reg}{\lambda}

\let\ab\allowbreak

%% file: symbols.tex
\newcount\K 
\def\blah{
    \K=0 \loop\ifnum\K<20 
    {\color[rgb]{0.8, 0.8, 0.8} blah blah blah blah blah blah blah blah blah blah} 
    \advance\K by1\repeat 
}

\usepackage{mleftright}
\mleftright

\newcommand{\kh}{two-layer Hopfield network}  
\newcommand{\kro}{hierarchical Hopfield network}

\newcommand{\Ns}{N_s}
\newcommand{\Nv}{N_{v_s}\times N_{v_c}}
\newcommand{\Nc}{N_c}

\newcommand{\xs}{x^s} \newcommand{\ts}{\tau_s} \newcommand{\gs}{g^s}
\newcommand{\xv}{x^v} \newcommand{\tv}{\tau_v} \newcommand{\gv}{g^v}
\newcommand{\xc}{x^c} \newcommand{\tc}{\tau_c} \newcommand{\gc}{g^c}

\newcommand{\xii}[2]{\xi^{(#1, #2)}}
\newcommand{\axii}[2]{\tilde{\xi}^{(#1, #2)}}

\DeclareMathOperator{\LN}{LayerNorm}
\DeclareMathOperator{\gelu}{GELU}
\DeclareMathOperator{\relu}{ReLU}
\DeclareMathOperator{\igelu}{\Phi_{\text{GELU}}}

\newcommand{\emixer}{Energy MetaFormer}
\newcommand{\vmixer}{VanillaMixer}
\newcommand{\pmixer}{ParaMixer}
\newcommand{\smixer}{SymMixer}
\newcommand{\psmixer}{Para/Sym-Mixer}
\newcommand{\asmixer}{AsymMixer}

\newcommand{\githubrepo}{\url{https://github.com/Toshihiro-Ota/paramixer}}

\newcommand{\blfootnote}[1]{
\renewcommand{\thefootnote}{\fnsymbol{footnote}}
{\protect\NoHyper\footnote[0]{#1}\protect\endNoHyper}
\renewcommand{\thefootnote}{\arabic{footnote}}
}

\newcommand{\eref}[1]{Eq.~(\ref{#1})}
\newcommand{\fref}[1]{Fig.~\ref{#1}}
\newcommand{\tref}[1]{Table~\ref{#1}}
\newcommand{\sref}[1]{Sec.~\ref{#1}}
\newcommand{\appenref}[1]{Appendix~\ref{#1}}
\renewcommand{\algref}[1]{Algorithm~\ref{#1}}

\newcommand{\dd}[2]{\frac{d #1}{d #2}}
\newcommand{\dpdp}[2]{\frac{\partial #1}{\partial #2}}

\newcommand{\set}[1]{\left\{ #1 \right\}}
\newcommand{\abs}[1]{\left\lvert #1 \right\rvert}
\newcommand{\norm}[1]{\left\lVert #1 \right\rVert}

%% file: abstract.tex
\begin{abstract}

Transformers have established themselves as the leading neural network model in natural language processing and are increasingly foundational in various domains.
In vision, the MLP-Mixer model has demonstrated competitive performance, suggesting that attention mechanisms might not be indispensable.
Inspired by this, recent research has explored replacing attention modules with other mechanisms, including those described by MetaFormers.
However, the theoretical framework for these models remains underdeveloped.
This paper proposes a novel perspective by integrating Krotov's hierarchical associative memory with MetaFormers, enabling a comprehensive representation of the entire Transformer block, encompassing token-/channel-mixing modules, layer normalization, and skip connections, as a single Hopfield network.
This approach yields a parallelized MLP-Mixer derived from a three-layer Hopfield network, which naturally incorporates symmetric token-/channel-mixing modules and layer normalization.
Empirical studies reveal that symmetric interaction matrices in the model hinder performance in image recognition tasks.
Introducing symmetry-breaking effects transitions the performance of the symmetric parallelized MLP-Mixer to that of the vanilla MLP-Mixer.
This indicates that during standard training, weight matrices of the vanilla MLP-Mixer spontaneously acquire a symmetry-breaking configuration, enhancing their effectiveness.
These findings offer insights into the intrinsic properties of Transformers and MLP-Mixers and their theoretical underpinnings, providing a robust framework for future model design and optimization.
\blfootnote{Author names are in alphabetical order.}

\end{abstract}

%% file: sec1.tex
\section{Introduction}

Transformers \cite{NIPS2017_3f5ee243}, widely recognized as the preeminent neural network model in natural language processing, are now solidifying their position as a foundational technology across various domains.
In the vision domain, the Vision Transformer (ViT) \cite{DosovitskiyICLR2021,TouvronICML2021} is increasingly replacing the traditional convolutional neural networks, achieving significant success.
Shortly after the advent of ViT, the MLP-Mixer model, which utilizes only a two-layer MLP instead of an attention module, was proposed \cite{tolstikhin2021mlp,melas2021you}, suggesting that the attention may not be indispensable in the vision domain.
This discovery catalyzed research into replacing attention modules with other mechanisms, leading to the recent proposal of a family of models abstracted as MetaFormers \cite{yu2022metaformer,yu2022metaformer2}.
Observations from MetaFormers indicate that the token mixer in the architecture can be highly flexible, yet there is no concrete guideline for designing an appropriate token mixer for specific tasks within the MetaFormer context.
Furthermore, while MetaFormers have generically exhibited good empirical performance, their theoretical treatment remains challenging and underdeveloped.

The Hopfield network is a classical associative memory model of single-layer recurrent neural network \cite{hopfield82,hopfield84}.
In this network, stored memories and their retrieval are well described by the attractors of an energy function and the dynamics converging to them.
For many Hopfield networks, it was assumed for years that neuron states and interaction matrices take discrete values, making it challenging to integrate them into the modern framework of deep neural networks that learn through backpropagation using differential methods.
Recently, Ramsauer et al.~proposed a modern Hopfield network with continuous states and interactions \cite{ramsauer2021hopfield,NEURIPS2020_da4902cb}.
They claimed that the update rule for neuron states in a modern Hopfield network with a specific energy function is essentially equivalent to the attention module, whereas their discussion was heuristic and somewhat ad hoc.
Around the same time, Krotov and Hopfield introduced a \kh ~equipped with Lagrangian functions that define the system \cite{krotov2020large}.
They demonstrated that the modern Hopfield network of Ramsauer et al.~can be realized as a special case within the framework of the \kh ~by choosing particular Lagrangian functions.
Subsequently, Krotov extended this \kh ~to a hierarchical model, successfully constructing a general $L$-layer hierarchical associative memory model \cite{krotov2021hierarchical}.
In the framework of these generalized Hopfield networks by Krotov and Hopfield, which we refer to as the \kro s, each model of Hopfield network is instantiated by determining a specific combination of Lagrangians.
Models derived in this manner thus inherently possess energy functions, with clear mathematical properties.
It has been shown that by adopting appropriate Lagrangians, modules such as attention, two-layer MLP, convolution, and pooling—all of which are used as token mixers in MetaFormers—can be derived \cite{krotov2020large,krotov2021hierarchical,tang2021remark} from the \kro.

Inspired by the fact that \kro s encompass a certain class of MetaFormers, recent research has begun to explore the fundamental properties of Transformers and MLP-Mixers (more generally, MetaFormers) through the lens of Hopfield networks \cite{hoover2024energy,ota2023imixer}.
In previous studies, the correspondence between \kro s and MetaFormer models indicated that only token-mixing modules such as attention and spatial MLP correspond to specific types of Hopfield networks.
In this paper, we propose a new model by introducing a novel perspective into Krotov's hierarchical associative memory, allowing the entire Transformer (MetaFormer) block, not just the token-mixing module, but also the channel-mixing module, layer normalization, and skip connection, to correspond exactly to a single Hopfield network.
This model, derived from a three-layer Hopfield network, naturally yields MetaFormers with parallelized token-/channel-mixing modules.
Based on this theoretical derivation, we focus on a specific combination of Lagrangians, presenting a parallelized MLP-Mixer, and examine its fundamental properties.
Thinking of it as a stack of associative memory models, interaction matrices of the proposed model exhibits a certain symmetry (symmetric parallelized MLP-Mixer, \emph{\smixer}), while empirical observations show that this symmetry actually hinders its performance as an image recognition model.
By introducing symmetry-breaking effects into the interaction weight matrices, we confirm that the performance of \smixer ~essentially transitions to that of the vanilla MLP-Mixer.
This suggests that during the normal training of the vanilla MLP-Mixer, the weight matrix spontaneously acquires a symmetry-breaking configuration through learning.

The rest of this paper is organized as follows.
After the list of related works, we provide a general background for our analysis, giving a brief overview of the \kh ~and showing the correspondence between a certain type of Hopfield network and the MLP-Mixer.
We then move on to discussing the hierarchical associative memory model, which is the main objective of this paper.
In \sref{sec:emixer}, we introduce a three-layer Hopfield network and study its basic properties as an associative memory model.
A prototype of the parallelized MetaFormer emerges from a specific configuration of neurons and a combination of Lagrangians.
In the subsequent section, we derive a class of MLP-Mixers with parallelized mixing layers as a stack of associative memory models.
Empirical studies reveal that the symmetric weights of mixing layers are indeed a constraint for image recognition tasks.
In \sref{sec:asmixer}, we demonstrate that, on the Hopfield model side, symmetry breaking resolves degeneracies of local minima of an energy function, and it plays a crucial role in ensuring performance on the Mixer model side.%
\footnote{The code for our experiments is available at \githubrepo .}

%% file: sec2.tex
\section{Related Work}

While it is widely believed that an attention mechanism is critical to the success of ViT, there have also been attention-free alternatives.
MLP-Mixer \cite{tolstikhin2021mlp, melas2021you, touvron2022resmlp, liu2022we} has shown that by simply replacing the attention mechanism of ViT with MLP, it is possible to achieve performance approaching that of ViT.
This discovery has stimulated a series of studies showing that a wide range of token-mixing mechanisms,
including pooling \cite{yu2022metaformer}, global filtering \cite{rao2021global}, recurrent layers \cite{tatsunami2022sequencer}, and graph neural networks \cite{han2022vision}, can be used as substitutes for the attention mechanism.
Wang et al.~\cite{wang2023riformer} demonstrate that the token mixer can even be removed.

The classical Hopfield network \cite{hopfield82,hopfield84} is known to suffer from limited memory capacity.
To address this issue, models with substantially higher memory capacities have been proposed \cite{NIPS2016_eaae339c,demircigil2017model,krotov2018dense}, but they require many-body interactions among neurons, which is biologically implausible in the brain.
Recently, Krotov and Hopfield developed more comprehensive associative memory models that consist of visible and hidden neurons with only two-body interactions between them \cite{krotov2020large,krotov2021hierarchical}.
Based on their findings, there has been a growing number of studies aiming to understand modern neural networks through the perspective of Hopfield networks
\cite{millidge2022universal,ota2023attention,ota2023learning,hoover2023memory,yampolskaya2023controlling,ambrogioni2023in,lucibello2024exponential}; see also \cite{krotov2023new}.

%% file: sec3.tex
\section{Background}

To clarify the notation, we provide here the derivation of the mixing layers in MLP-Mixer \cite{tolstikhin2021mlp} from a continuous Hopfield network.

\subsection{Overview of the \kh}
\label{sec:Overview}

\begin{figure}[t]
    \begin{minipage}[b]{0.49\linewidth}
        \centering
        \includegraphics[keepaspectratio, scale=0.2]{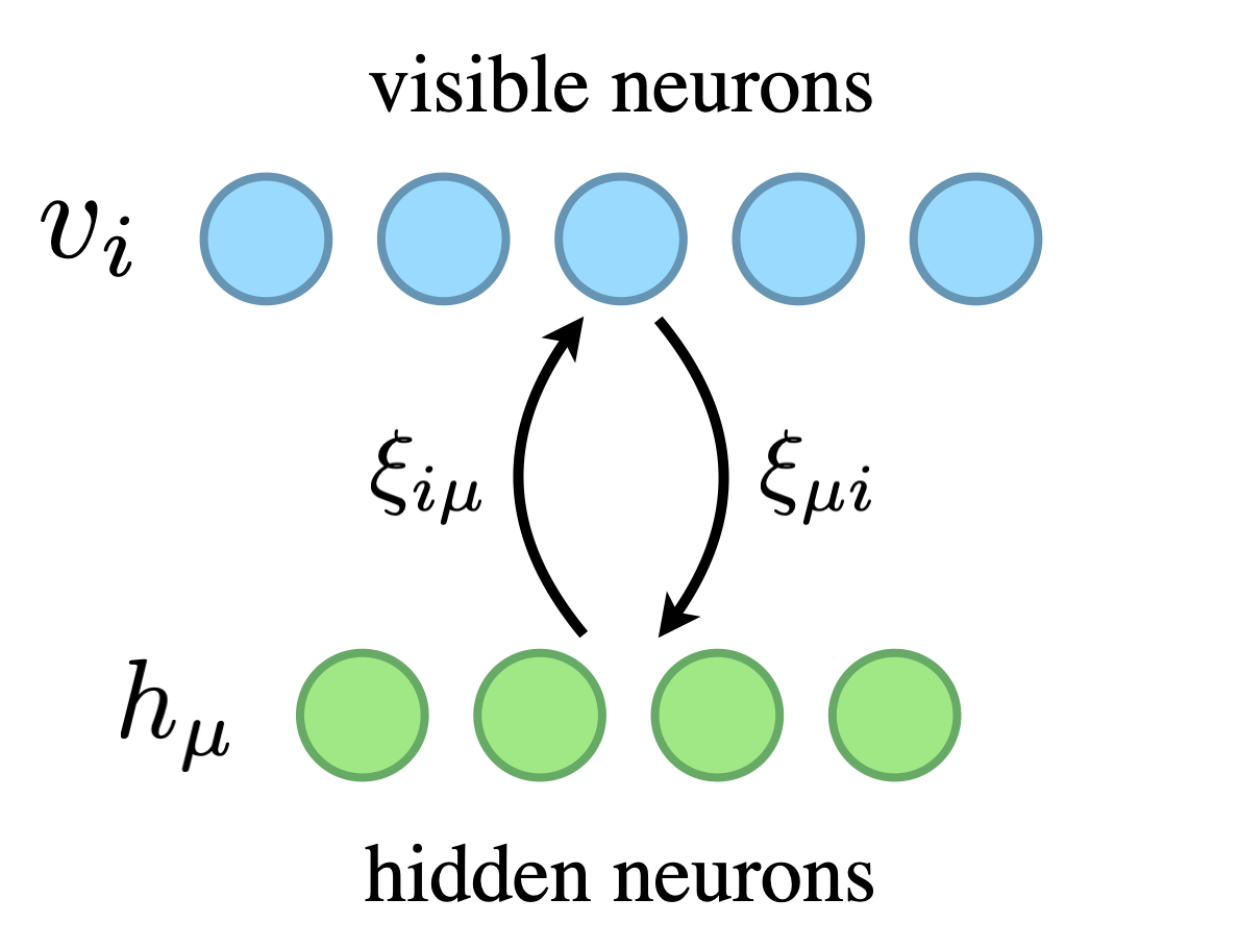}
        \subcaption{}  
        \label{fig:neurons}
    \end{minipage}
    \hspace{-1em}
    \begin{minipage}[b]{0.49\linewidth}
        \centering
        \includegraphics[keepaspectratio, scale=0.26]{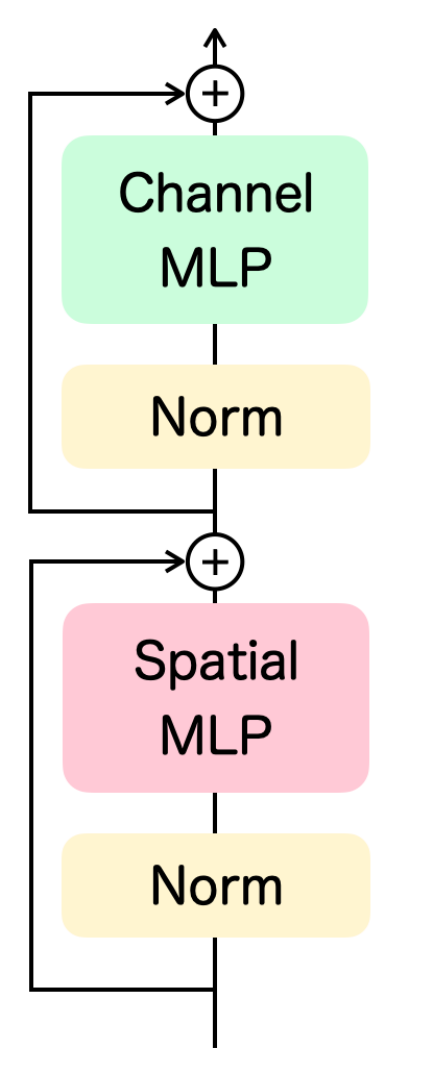}
        \subcaption{}  
        \label{fig:vanillamixer}
    \end{minipage}
    \caption{
        The \kh.
        (a) Visible and hidden neurons interact each other and the layers form a bipartite graph.
        (b) Each of the token- (spatial) and channel-mixing block in the mixing layer of MLP-Mixer can be realized as an instance with the corresponding Lagrangians.
        }
\end{figure}

Let us first briefly review the \kh ~proposed in \cite{krotov2020large}.
In this system, the dynamical variables are composed of $N_v$ visible neurons and $N_h$ hidden neurons both continuous,
\begin{equation}
    v(t) \in \R^{N_v},  \quad  h(t) \in \R^{N_h},
\end{equation}
where the argument $t$ can be thought of as ``time''.
The interaction matrices between them,
\begin{equation}
    \xi \in \R^{N_h \times N_v},  \quad  \xi' \in \R^{N_v \times N_h},
\end{equation}
are basically supposed to be symmetric: $\xi' = \xi^{\top}$, see \fref{fig:neurons}.
With the relaxing time constants of the two groups of neurons $\tau_v$ and $\tau_h$, the dynamics of the system is described by the following differential equations,
\begin{align}
    \tau_v \dd{v_i(t)}{t}
        &= \sum_{\mu=1}^{N_h} \xi_{i \mu} f_\mu(h(t)) - v_i(t),  \\
    \tau_h \dd{h_\mu(t)}{t}
        &= \sum_{i=1}^{N_v} \xi_{\mu i} g_i(v(t)) - h_\mu(t),  \label{eq:heq}
\end{align}
where the activation functions $f$ and $g$ are determined through Lagrangians $L_h: \R^{N_h}\to \R$ and $L_v: \R^{N_v}\to \R$, such that
\begin{equation}
    f_\mu(h) = \dpdp{L_h(h)}{h_\mu},  \quad  g_i(v) = \dpdp{L_v(v)}{v_i}.
\end{equation}

The canonical energy function for this system is given by
\begin{equation}
    E(v, h)
        = \sum_{i=1}^{N_v} v_i g_i(v) - L_v(v)  
        + \sum_{\mu=1}^{N_h} h_\mu f_\mu(h) - L_h(h)
        - \sum_{\mu, i} f_\mu \xi_{\mu i} g_i.
\end{equation}
One can easily find that this energy function monotonically decreases along the trajectory of the dynamical equations to define an associative memory model,
\begin{equation}
    \dd{E(v(t), h(t))}{t} \leq 0,
\end{equation}
provided that the Hessians of the Lagrangians are positive semi-definite.
In addition to this, if the overall energy function is bounded from below, the trajectory is guaranteed to converge to a fixed-point attractor state, which corresponds to one of the local minima of the energy function.
Such fixed points and the process of convergence are thought of as associative memories and memory retrieval of an associative memory model.
The formulation of neural networks in terms of Lagrangians and the associated energy functions enables us to easily experiment with different choices of the activation functions and different architectural arrangements of neurons.

\subsection{Mixing layer as an associative memory model}
\label{sec:Mixer}

Suppose we have a fixed interaction matrix $\xi_{\mu i}$, then the system is defined by the choice of Lagrangians $L_h$ and $L_v$.
Tang and Kopp demonstrated that the specific choice of Lagrangians called ``model C'' in \cite{krotov2020large} essentially reproduces the mixing layers in the MLP-Mixer \cite{tang2021remark},
which is given by the following Lagrangians:
\begin{equation}
    L_h(h) = \sum_\mu F(h_\mu),  \quad  L_v(v) = \sqrt{ \sum_i (v_i - \bar{v})^2 },
    \label{eq:Lmixer}
\end{equation}
where $F$ will be specified below, and $\bar{v} = \sum_i v_i / N_v$.
For these Lagrangians, the activation functions are%
\footnote{$g_i$ can include learnable parameters such as $\bm{\alpha}$ and $\bm{\beta}$ as in \cite{ba2016layer} by slightly modifying the Lagrangian $L_v$, see \appenref{appendix:formulae}.}
\begin{align}
    f_\mu(h)
        &= \dpdp{L_h}{h_\mu}
        = F'(h_\mu),  \\
    g_i(v)
        &= \dpdp{L_v}{v_i}
        = \frac{v_i - \bar{v}}{\sqrt{\sum_j (v_j - \bar{v})^2}}
        = \LN(v)_i.
\end{align}

We now consider the adiabatic limit, $\tau_v \gg \tau_h$, which means that the dynamics of the hidden neurons is much faster than that of the visible neurons,
i.e., we can take $\tau_h \to 0$:
\begin{equation}
    \text{\eref{eq:heq}}
    \quad  \rightsquigarrow  \quad
    h_\mu(t) = \sum_{i=1}^{N_v} \xi_{\mu i} \LN(v(t))_i.
\end{equation}
By substituting this expression into the other dynamical equation, we find
\begin{equation}
    \tau_v \dd{v_i(t)}{t}
        = \sum_\mu \xi_{i \mu} F'\bigg( \sum_j \xi_{\mu j} \LN(v(t))_j \bigg) - \alpha v_i(t).
\end{equation}
Notice that we can put an arbitrary coefficient $\alpha$, which can be even zero, in front of the decay term,
since for this choice of Lagrangian $L_v$ its Hessian has a zero mode:
\begin{equation}
    \sum_j M_{ij}(v_j - \bar{v}) = 0,
    \quad
    M_{ij} := \frac{\partial^2 L_v}{\partial v_i \partial v_j}.
\end{equation}

If we take $\alpha=0$ and discretize the differential equation by taking $\Delta t = \tau_v$, then we obtain the update rule for the visible neurons as
\begin{equation}
    v_i(t+1) = v_i(t) + \sum_\mu \xi_{i \mu} \, \sigma\bigg( \sum_j \xi_{\mu j} \LN(v(t))_j \bigg),
    \label{eq:vanillamixer}
\end{equation}
where we defined $\sigma := F'$.
If it is chosen as $\sigma = \gelu$, this update rule is identified with the token- and channel-mixing blocks in the mixing layers discussed in \cite{tolstikhin2021mlp}, as in \fref{fig:vanillamixer}.

%% file: sec4.tex
\section{Hierarchical Associative Memory Model}
\label{sec:emixer}

Along the line of \cite{krotov2020large}, Krotov further extended the \kh ~to a hierarchical associative memory model with multiple hidden layers \cite{krotov2021hierarchical}, which we refer to as the \kro.
Based on observations in \sref{sec:Mixer} and Krotov's extension, in the subsequent sections we will study a type of MetaFormers through the lens of associative memory models and their variants.

First, in this section we consider the hierarchical extension of the discussion of the previous section.
We derive a prototype of the parallelized MetaFormers via a different viewpoint from the original formulation of the \kro, and demonstrate its basic properties as an associative memory model.

\subsection{\emixer}

\begin{figure}[t]
    \begin{minipage}[b]{0.3\linewidth}
        \centering
        \includegraphics[keepaspectratio, scale=0.18]{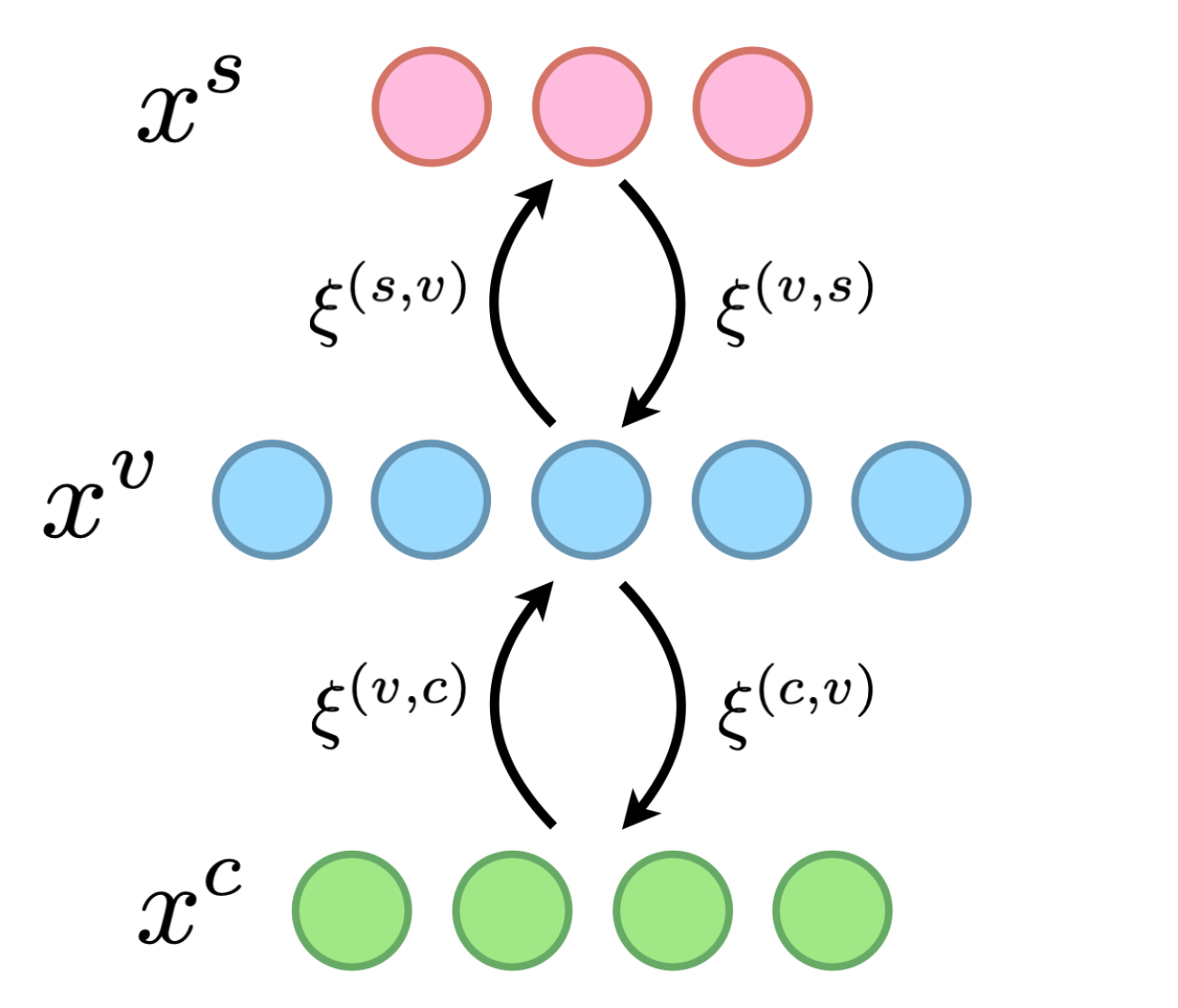}
        \subcaption{Hierarchical Hopfield network}
        \label{fig:hierarchical}
    \end{minipage}
    \begin{minipage}[b]{0.39\linewidth}
        \centering
        \includegraphics[keepaspectratio, scale=0.2]{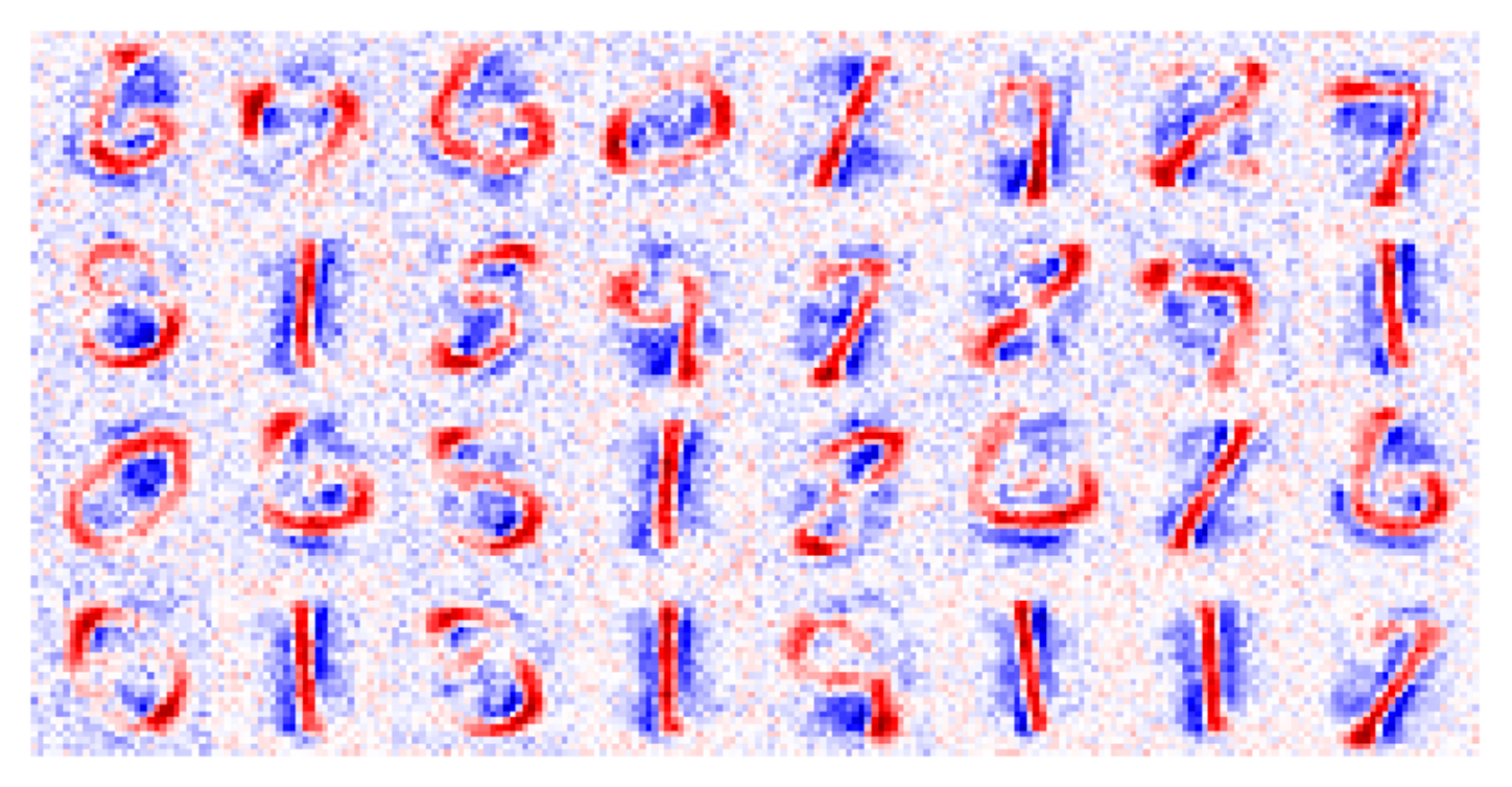}
        \vspace{.1em}
        \subcaption{Receptive fields}
        \label{fig:em-recep}
    \end{minipage}
    \begin{minipage}[b]{0.3\linewidth}
        \centering
        \includegraphics[keepaspectratio, scale=0.22]{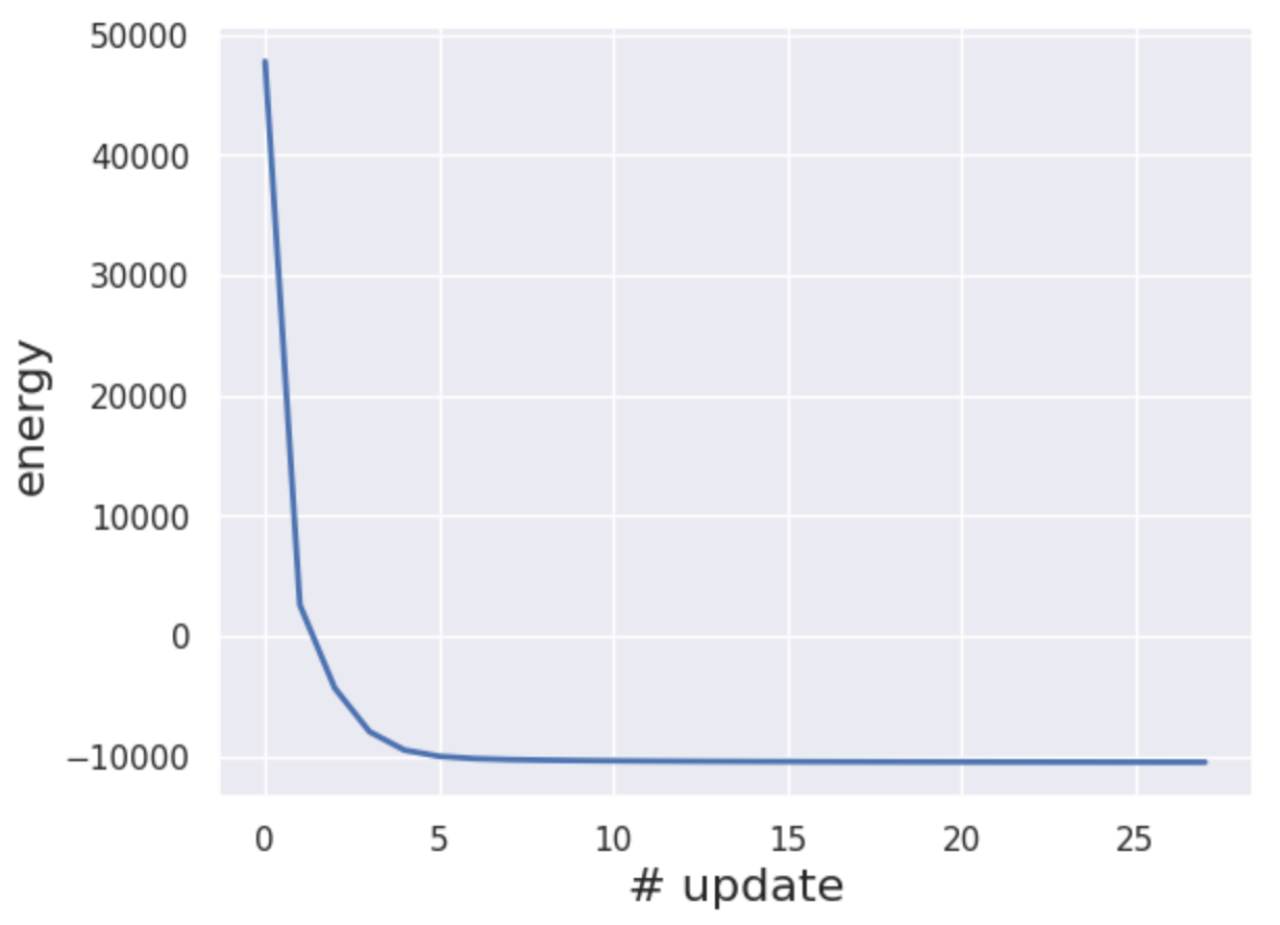}
        \subcaption{Energy descent}
        \label{fig:em-energy}
    \end{minipage}
    \caption{\emixer.}
\end{figure}

As a simple extension in the context of \kro, we consider a structure with three layers: one visible layer with $N_v$ neurons and two hidden layers with $\Ns$ and $\Nc$ neurons each.
The crucial point is that the set of the visible neurons lies in between the two hidden layers as depicted in \fref{fig:hierarchical}, unlike the original configuration discussed in \cite{krotov2021hierarchical}.
The dynamics of this system is then described by the following differential equations,
\begin{align}
    \ts \dd{\xs_\alpha(t)}{t}
        & = \sum_{i=1}^{N_v} \xii{s}{v}_{\alpha i} \gv_i(\xv(t)) - \xs_\alpha(t),  \\
    \tv \dd{\xv_i(t)}{t}
        & = \sum_{\alpha=1}^{\Ns} \xii{v}{s}_{i\alpha} \gs_\alpha(\xs(t))
          + \sum_{\beta=1}^{\Nc} \xii{v}{c}_{i\beta} \gc_\beta(\xc(t))
          - \xv_i(t), \\
    \tc \dd{\xc_\beta(t)}{t}
        & = \sum_{i=1}^{N_v} \xii{c}{v}_{\beta i} \gv_i(\xv(t)) - \xc_\beta(t),
\end{align}
where $\xii{A}{B}$ denotes the interaction from $B$ neurons to $A$ neurons: $\xii{A}{B}\in \R^{N_A\times N_B}$.
The activation functions are again determined through the corresponding Lagrangians,
\begin{equation}
    \gs_\alpha(\xs) = \dpdp{L^s(\xs)}{\xs_\alpha},
    \quad
    \gv_i(\xv) = \dpdp{L^v(\xv)}{\xv_i},
    \quad
    \gc_\beta(\xc) = \dpdp{L^c(\xc)}{\xc_{\beta}}.
\end{equation}

In this system, the canonical energy function is given by \cite{krotov2021hierarchical}
\begin{equation}
    E(\xs, \xv, \xc)
    = \sum_{A=s,v,c} \left( \left( x^A \right)^\top g^A - L^A \right)  -  \sum_{A=s,v} \left( g^{A+1} \right)^\top \xii{A+1}{A} g^A,
    \label{eq:energy-kro}
\end{equation}
where $s+1 := v$ and $v+1 := c$, and the arguments for $g^A$ and $L^A$ are omitted for simplicity.
We have moved on to the matrix notation for brevity, and all the products here after imply matrix multiplication.
Under the symmetric condition on the interaction matrices,  
\begin{equation}
    \xii{v}{s} = \left( \xii{s}{v} \right)^\top,  \quad  \xii{v}{c} = \left( \xii{c}{v} \right)^\top,
\end{equation}
and ensuring the Hessians of the Lagrangians are positive semi-definite, the above dynamical equations confirm that
\begin{equation}
    \dd{E(\xs(t), \xv(t), \xc(t))}{t} \leq 0.
\end{equation}
This fact demonstrates that we have a well-defined Lyapunov function for this system, which monotonically decreases along the trajectory of the dynamical equations.

We now take the specific Lagrangian for the visible neurons as
\begin{equation}
    L^v(\xv) = \sqrt{ \sum_{i=1}^{N_v} (\xv_i - \bar{v})^2 },
    \quad
    \bar{v} = \frac{1}{N_v} \sum_i \xv_i.
\end{equation}
Then, the dynamical equations become
\begin{align}
    \ts \dd{\xs(t)}{t}
        & = \xii{s}{v} \LN(\xv(t)) - \xs(t),  \nonumber  \\
    \tv \dd{\xv(t)}{t}
        & = \xii{v}{s} \gs(\xs(t)) + \xii{v}{c} \gc(\xc(t)) - \xv(t),  \label{eq:emixer}  \\
    \tc \dd{\xc(t)}{t}
        & = \xii{c}{v} \LN(\xv(t)) - \xc(t).  \nonumber
\end{align}
It is not necessary to specify the activation functions $\gs$ and $\gc$ at this stage.
Following the discussion in \sref{sec:Mixer}, the discrete update rule for $\xv$ neurons will lead to a new type of MLP-Mixer model, which we will discuss in the next section.
The flexible activation functions $\gs$ and $\gc$ suggest that the system may form a class of MetaFormers in turn.
Thus, these dynamical equations represent a hierarchical associative memory model, serving as a continuous MetaFormer endowed with an energy function.
This fact motivates us to refer to this system as the \emph{\emixer}.

\subsection{Numerical demonstration}

Before getting into the discussion of deep neural network, it is informative to demonstrate the functionality of the \emixer ~as an associative memory.
To gain intuition, in this subsection we instantiate a model by taking a specific combination of Lagrangians and train the model on a denoising task using Eqs.~(\ref{eq:emixer}).
For the numerical demonstration, we utilize the HAMUX framework \cite{hoover2022universal}\footnote{\url{https://github.com/bhoov/barebones-hamux}} to implement the \emixer.

Here, we set the dimensions of each layer to $N_v=784$ and $\Ns=\Nc=900$, and take the hidden Lagrangians as $\frac12 \max(x,0)^2$, which results in the $\relu$ activation function: $\gs=\gc=\relu$.
The energy function for this system is determined by the Lagrangians,
\begin{align}
    E(\xs, \xv, \xc)
    &= {\xv}^{\top} \LN(\xv)  -  \sqrt{ \sum_i (\xv_i - \bar{v})^2 }  \nonumber  \\
    &+ \sum_{A=s,c} \left( \left( x^A \right)^\top \relu(x^A)  -  \frac12 \max(x^A,0)^2 \right)  \nonumber  \\
    &- \LN(\xv)^\top \xii{v}{s} \relu(\xs)  -  \relu(\xc)^\top \xii{c}{v} \LN(\xv)
    .
    \label{eq:energy-emixer}
\end{align}
The model is trained on 60,000 training samples from the MNIST dataset \cite{deng2012mnist}.
The mini-batch size is set to 512.
We add Gaussian noise to the training batches before inputting them into the model.
The input (initial state of visible neurons) evolves by the dynamical equations (\ref{eq:emixer}) and the model eventually outputs the state at one of the minima of the energy function \eref{eq:energy-emixer}.
We compute the mean-squared error as the objective function between the outputs and the true images.
The model is trained for 100 epochs with the Adam \cite{KingBa15} optimizer (using Optax \cite{deepmind2020jax} defaults: $\beta_1=0.9$ and $\beta_2=0.999$) at a constant learning rate of $10^{-4}$.

\Figref{fig:em-recep} depicts the receptive fields (a part of the weight vectors $\xii{s}{v}_\alpha$ and $\xii{c}{v}_\beta$) of the \emixer ~learned by the training, which roughly correspond to the minima of the energy function adapted to the MNIST training samples.
We can also verify the energy descent of the model.
Randomly initialized neuron states of the trained model evolve according to the dynamical equations (\ref{eq:emixer}) to converge to one of the minima of the energy function, as illustrated in \fref{fig:em-energy}.
These results demonstrate that the \emixer ~has successfully learned to embed the training dataset into the network (weight matrices) and functions effectively as an associative memory model.

%% file: sec5.tex
\section{Parallelized MLP-Mixer}

In this section, we advance the discussions from the previous section to derive a parallelized MLP-Mixer from the \kro.
We also perform experiments to explore the visual recognition capabilities of the proposed models, keeping in mind that they are conceived as a stack of associative memory models.

\subsection{Model}

\begin{figure}[t]
    \begin{minipage}[b]{0.495\linewidth}
        \centering
        \includegraphics[keepaspectratio, scale=0.22]{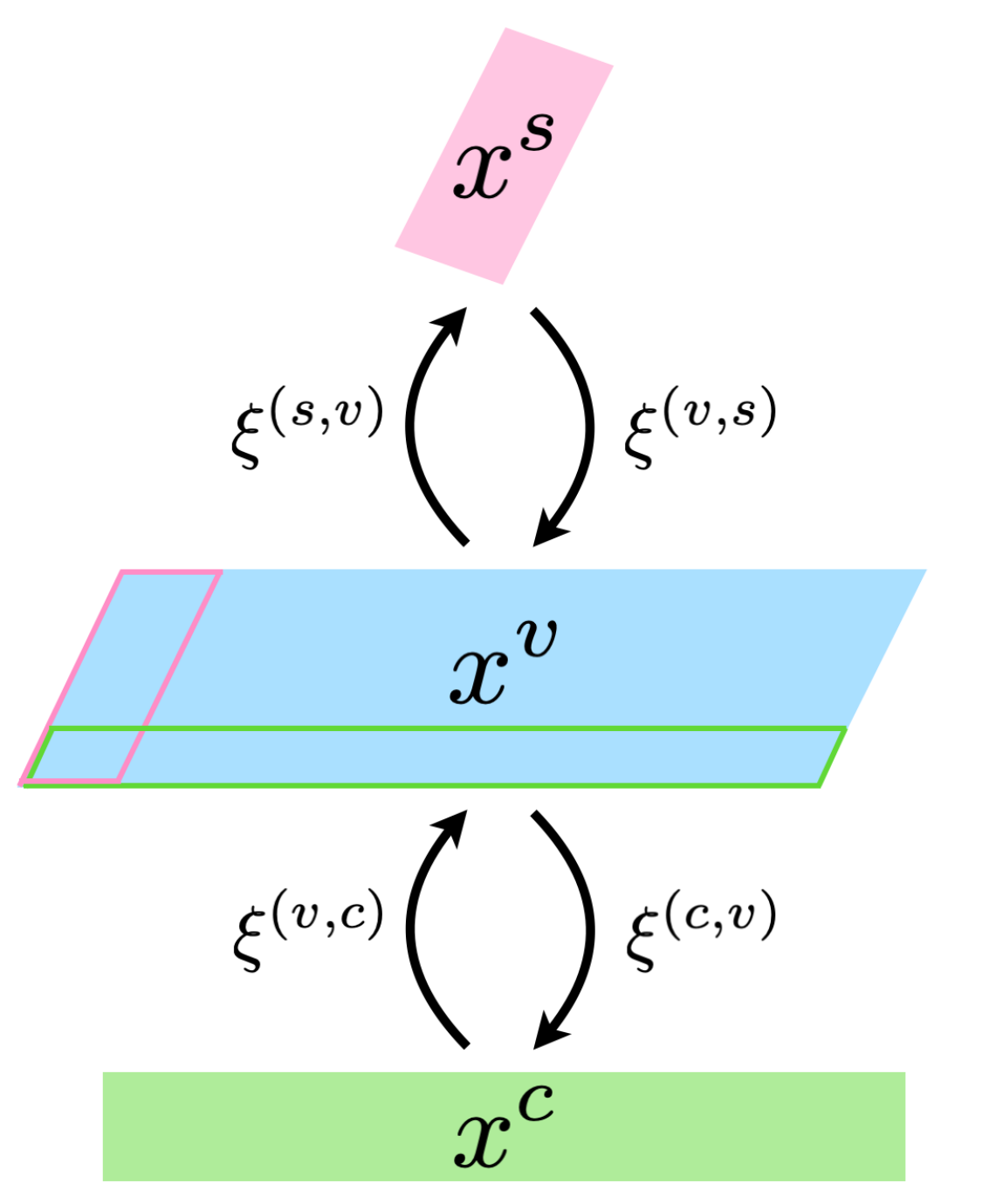}
        \subcaption{}
        \label{fig:hierarchical2}
    \end{minipage}
    \hspace{-1em}
    \begin{minipage}[b]{0.495\linewidth}
        \centering
        \includegraphics[keepaspectratio, scale=0.3]{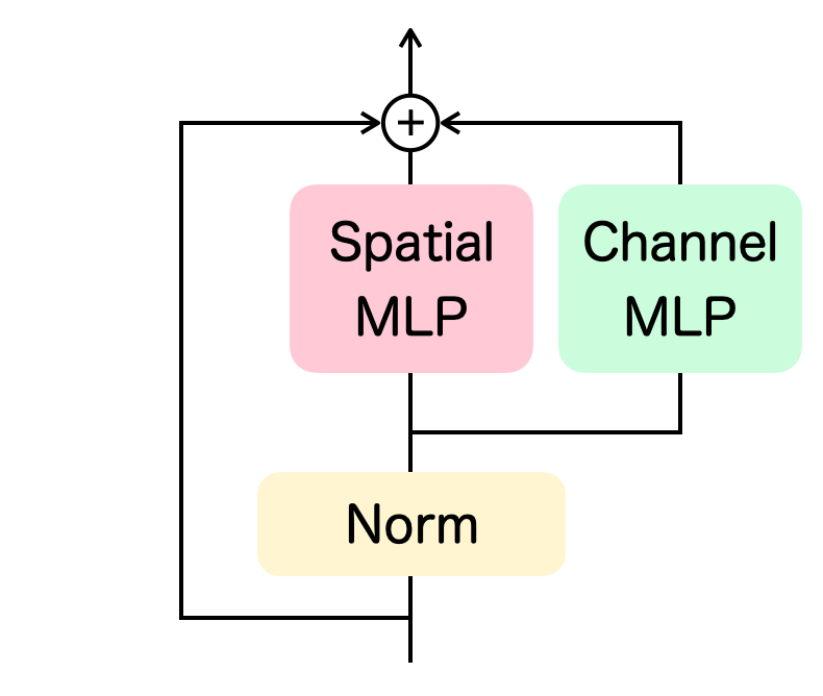}
        \subcaption{}
        \label{fig:paramixer}
    \end{minipage}
    \caption{
        Parallelized MLP-Mixer.
        (a) Three-layer \kro ~with 2d structure for visible neurons. Hidden layers interact with the visibles along the relatively perpendicular directions.
        (b) The whole mixing layer composed of parallelized token- and channel-mixing modules is identified with an update rule of a \kro.}
\end{figure}

To derive the model, we essentially follow the three-layer \kro ~setup discussed in the previous section.
A key point here is to introduce a two-dimensional structure into the layer of visible neurons, as shown in \fref{fig:hierarchical2}.
The set of visible neurons can now be considered as a matrix rather than a vector, $\xv\in\R^{\Nv},~ N_v=\Nv$, and interacts with the hidden neurons $\xs$ and $\xc$ along relatively perpendicular directions.

We continue to study the system \eref{eq:emixer}, while it should be noted that $\xv$ has two indices, e.g.,
\begin{align}
    \left(\xii{s}{v}\LN(\xv)\right)_\alpha
        &= \sum_{i,I}\xii{s}{v}_{\alpha iI}\LN(\xv)_{iI}  \nonumber  \\
        &= \sum_{i,I}\xii{s}{v}_{\alpha iI} \frac{\xv_{iI} - \bar{v}}{\sqrt{\sum_{j,J} (\xv_{jJ} - \bar{v})^2}},
    \label{eq:two-indices}
\end{align}
etc, where $i=1,\dots,N_{v_s}$ and $I=1,\dots,N_{v_c}$.

We now consider the adiabatic limit, $\tv \gg \ts,\, \tc$, then the dynamical equations for $\xs$ and $\xc$ neurons are reduced to
\begin{equation}
    \xs(t) = \xii{s}{v} \LN(\xv(t)),
    \quad
    \xc(t) = \xii{c}{v} \LN(\xv(t)).
    \label{eq:sc-red}
\end{equation}
By substituting these expressions to the other one and by following the same discussion as in \sref{sec:Mixer}, the differential equation for $\xv$ eventually becomes the update rule for the visible neurons:
\begin{equation}
    \xv(t+1)
    = \xv(t)
    + \xii{v}{s} \gs\left( \xii{s}{v} \LN(\xv(t)) \right)
    + \xii{v}{c} \gc\left( \xii{c}{v} \LN(\xv(t)) \right).
\end{equation}

Furthermore, as for the comparison with the vanilla MLP-Mixer (or more generally, MetaFormers), writing symmetric matrices
\begin{align}  
    W_2 = W_1^\top;  \quad  W_1 \in \R^{\Ns \times N_{v_s}},  \quad  W_2 \in \R^{N_{v_s} \times \Ns},  \label{eq:symweight1}  \\
    W_4 = W_3^\top;  \quad  W_3 \in \R^{N_{v_c} \times \Nc},  \quad  W_4 \in \R^{\Nc \times N_{v_c}},  \label{eq:symweight2}
\end{align}
we truncate the interaction matrices between the visible and the hiddens as
\begin{align}
    \xii{s}{v} &= (\xii{s}{v}_{\alpha iI}) = (W_1)_{\alpha i},  \\
    \xii{c}{v} &= (\xii{c}{v}_{\beta iI})  = (W_4)_{\beta I}.
\end{align}
With these expressions, we obtain the update rule for the visible neurons as shown in \fref{fig:paramixer},
\begin{equation}
    \xv(t+1)
    = \xv(t)
    + W_2 \gs\left( W_1 \LN(\xv(t)) \right)
    + \left(\gc\left( \LN(\xv(t)) W_3 \right)\right) W_4.
    \label{eq:symmixer}
\end{equation}
Each term on the right-hand side reads the skip-connection, the token-mixing, and the channel-mixing modules, respectively.
The activation functions $\gs$ and $\gc$ are not specified at this stage.
In practice, we set them to $\gs = \gc = \gelu$ in experiments to compare the proposed models with the vanilla MLP-Mixer.

Our single update rule (\ref{eq:symmixer}) contains all the components of the mixing layer: token- and channel-mixing modules, layer normalization, and skip-connection, unlike \eref{eq:vanillamixer} in \sref{sec:Mixer}.
The two main differences from an ordinary mixing layer arise from our construction via the correspondence with the \kro : the layer normalization is simultaneously applied along both token and channel axes as in \eref{eq:two-indices},
and the layer processes features for token- and channel-mixing modules in a parallel manner.%
\footnote{%
    Recently, a parallel Transformer block is occasionally adopted for some notable models (e.g., \cite{chowdhery22palm,zhong2022a}), and it shows comparable performance empirically.
}
A model consisting of stacked associative memory models \eref{eq:symmixer} as mixing layers becomes an MLP-Mixer model composed of parallelized token- and channel-mixing modules.

\subsection{Experiments}
\label{sec:experiments}

We incorporate the update rule \eref{eq:symmixer} into the mixing layers, which results in a parallelized MLP-Mixer as a stack of associative memory models.
To investigate the trainability of the proposed model and the implications from the correspondence with the Hopfield network side, we conduct several empirical studies to compare our model with the vanilla MLP-Mixer (\vmixer).
The statistics of the results are obtained from ten trials with random initialization across all the experiments in this subsection.
The experimental details are provided in \appenref{appendix:experiments}.

\textbf{Network architecture.}
We consider two cases of the model:
parallelized MLP-Mixer with symmetric weights as in Eqs.~(\ref{eq:symweight1}) and (\ref{eq:symweight2}), which we refer to as \emph{\smixer}, and parallelized MLP-Mixer without weight constraint for comparison, which we simply refer to as \emph{\pmixer}.
We incorporate the proposed module \eref{eq:symmixer} with the mixing layers, and keep the rest of the networks exactly identical to \vmixer.
From this construction, \psmixer ~actually do not have additional hyperparameters compared to \vmixer.
The main difference between the proposed models and the ordinary MLP-Mixer is the normalization part within the mixing layers.
As already mentioned, the layer normalization is applied over both the token and channel axes.
This stems from the symmetry of the two directions inherent in the visible neurons as in \fref{fig:hierarchical2} and is thus natural from the correspondence with the Hopfield network side.
For control experiments, we implement this symmetric layer normalization for all the models, and examine some ablation studies below.
We utilize PyTorch Image Models \texttt{timm} \cite{rw2019timm} to implement the models in all the experiments in this section.

\textbf{Trainability.}
To study the trainability aspect of \psmixer, we perform scratch training of the models on a classification task with CIFAR-10/100 \cite{cifar}.
The CIFAR-10/100 datasets each consist of 60,000 natural images of size $32\times 32$.
The ground-truth object category labels are attached to each image, and the number of categories is 10/100, with 6,000/600 images per class.
There are 50,000 training images and 10,000 test images.
For the training setting, we basically follow the previous works \cite{TouvronICML2021,hou2022vision}.
The images are resized to $224\times 224$, and the AdamW \cite{loshchilovdecoupled} optimizer is used.
We set the base learning rate to $\frac{\text{batch size}}{512}\times 5\times 10^{-4}$, and the mini-batch size is 384.
As regularization methods, we employ label smoothing \cite{Szegedy_2016_CVPR} and stochastic depth \cite{huang2016deep}.
For data augmentation, we apply cutout \cite{devries2017improved}, cutmix \cite{Yun_2019_ICCV}, mixup \cite{zhangmixup}, random erasing \cite{zhong2020random}, and randaugment \cite{cubuk2020randaugment}.
For more details, see \appenref{appendix:training}.
We train all the Mixer models under the exact same training configuration for fair comparison.
All the trainings are conducted with four V100 GPUs.

\begin{table}[t]
    \centering
    \caption{Top-1 accuracy (\%) of the Mixer models.}
    \label{tab:result-cifar}
    \begin{tabular}{lcc}
        \toprule
            Model   &  CIFAR-10                          &  CIFAR-100                         \\
        \midrule
            \vmixer &  $89.61$ {\scriptsize $\pm 0.22$}  &  $69.13$ {\scriptsize $\pm 0.52$}  \\
            \pmixer &  $89.80$ {\scriptsize $\pm 0.29$}  &  $69.05$ {\scriptsize $\pm 0.53$}  \\
            \smixer &  $77.86$ {\scriptsize $\pm 0.53$}  &  $53.19$ {\scriptsize $\pm 0.21$}  \\
        \bottomrule
    \end{tabular}
\end{table}

\tref{tab:result-cifar} shows the results.
The performance of \pmixer ~is competitive with \vmixer, whereas \smixer ~exhibits significant performance drops.
This is quite counterintuitive since the difference between \pmixer ~and \smixer ~is only the symmetry constraint on the weight matrices.
From previous studies \cite{hu2019exploring,yang2022transformers}, it may be expected that many modern neural network architectures exhibit robust performance under certain symmetric constraints on weights.
The enormous performance drop from \pmixer ~to \smixer ~suggests that symmetric weights and their symmetry breaking have an important effect on the capability of mixing layers.
We will investigate this aspect in the next section.

\begin{figure}[t]
    \begin{minipage}[b]{0.495\linewidth}
        \centering
        \includegraphics[keepaspectratio, scale=0.28]{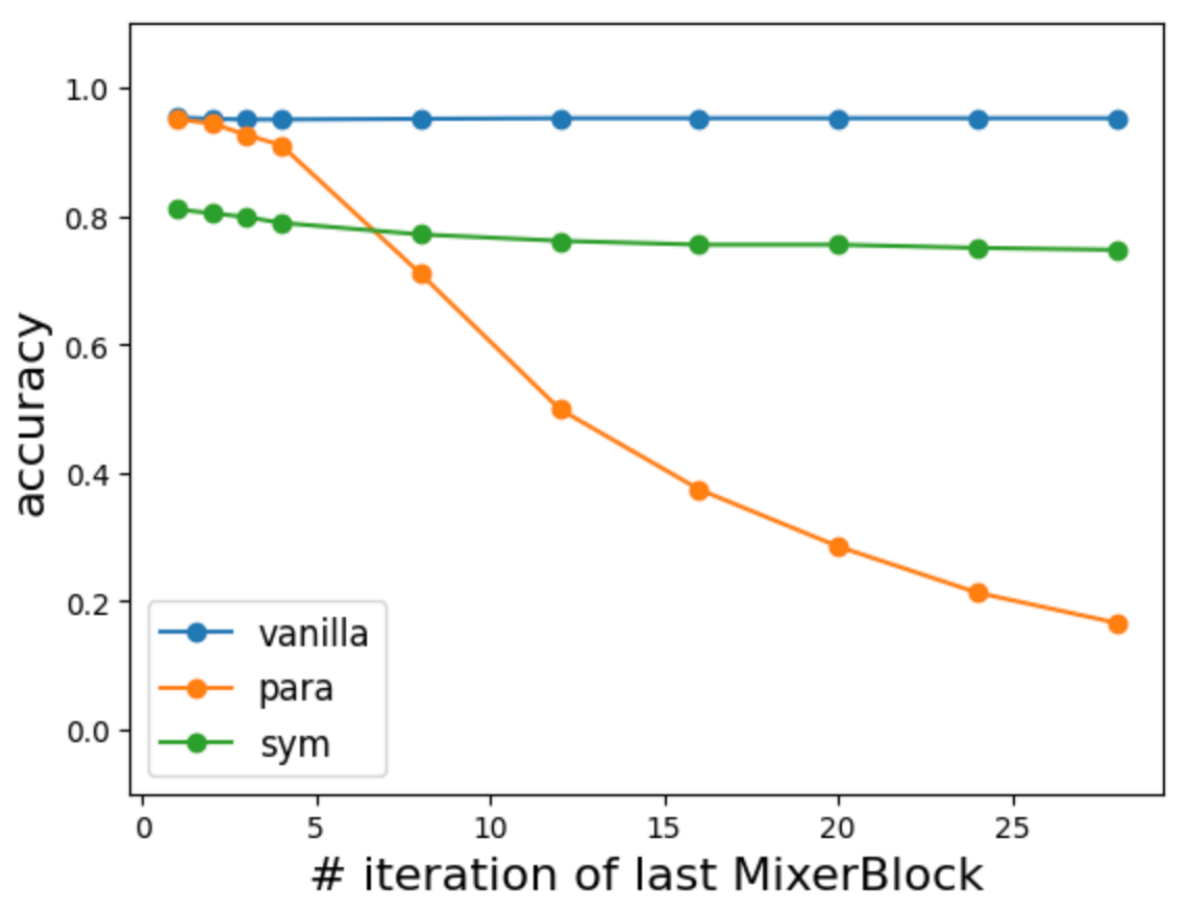}
        \subcaption{Train samples}
        \label{fig:acc-fni-train}
    \end{minipage}
    \begin{minipage}[b]{0.495\linewidth}
        \centering
        \includegraphics[keepaspectratio, scale=0.28]{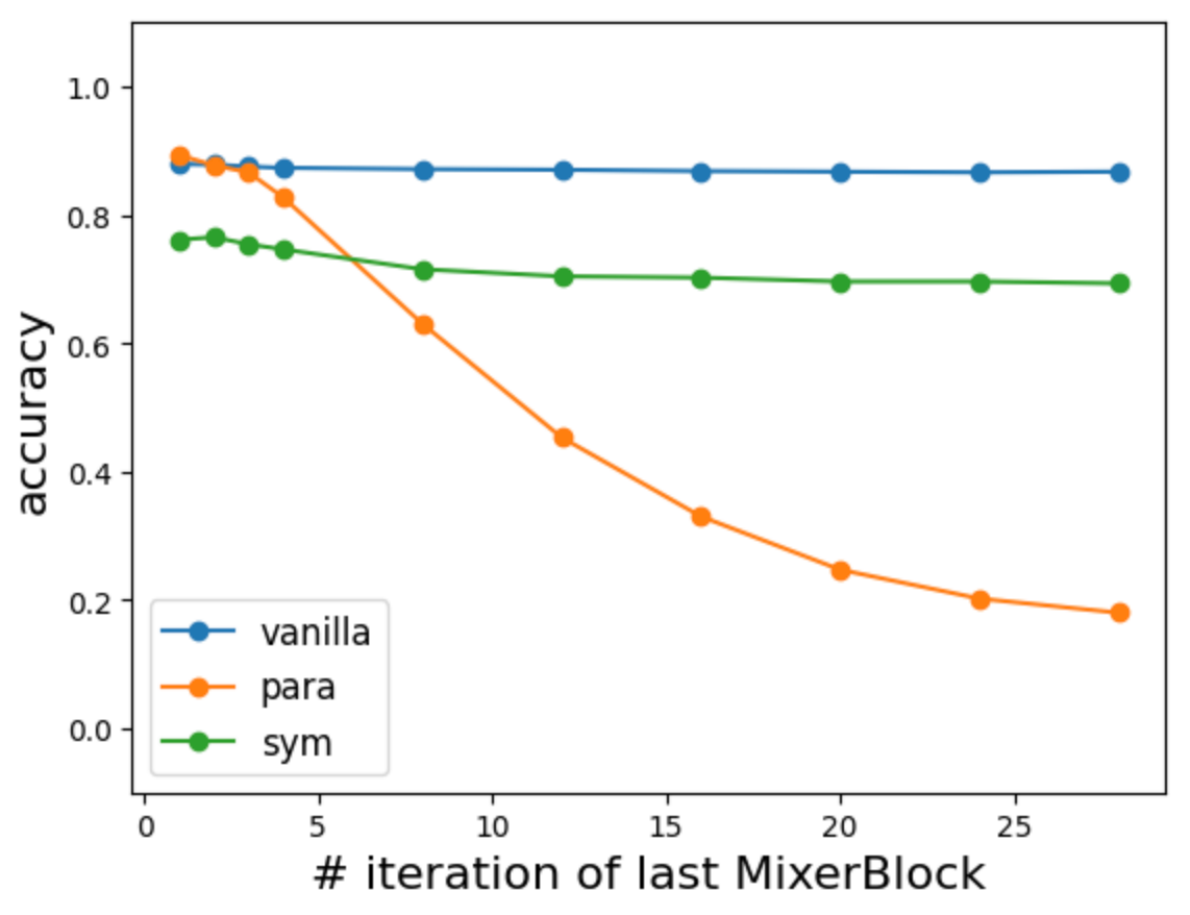}
        \subcaption{Test samples}
        \label{fig:acc-fni-test}
    \end{minipage}
    \caption{Accuracy vs number of iterations of the last mixing layer.}
    \label{fig:acc-fni}
\end{figure}

\textbf{Iterative mixing layers.}
Since the proposed mixing layer \eref{eq:symmixer} is viewed as a lump of an associative memory model, it is natural to consider iterative updates of neurons.
We use the trained Mixer models for CIFAR-10 and iteratively update the neuron states of the last mixing layer, which is regarded as the linear classifier of the network.
The results are shown in \fref{fig:acc-fni}.
The inputs are 1,000 train and test samples of CIFAR-10 chosen at random.
From these results, we observe that the classification performance of \pmixer ~significantly degrades as the number of iterations of the mixing layer increases, while \smixer ~maintains its performance over long iterations.
This suggests that \smixer ~stores associative memories of features during training and that the symmetric weights can properly retrieve them in the inference phase.
We have also plotted the results for \vmixer ~for comparison, but it is difficult to make any sense since the mixing layers of \vmixer ~have nothing to do with the Hopfield network side.

\begin{table}[ht]
    \centering
    \caption{
        Top-1 accuracy (\%) of the Mixer models trained with CIFAR-10 from scratch.
        All the mixing layers of the models are iteratively applied.
        }
    \label{tab:iter-cifar}
    \begin{tabular}{lccc}
        \toprule
            \# iteration &  \vmixer                           &  \pmixer                           &  \smixer                           \\
        \midrule
            1 (baseline) &  $89.61$ {\scriptsize $\pm 0.22$}  &  $89.80$ {\scriptsize $\pm 0.29$}  &  $77.86$ {\scriptsize $\pm 0.53$}  \\
            2            &  $90.38$ {\scriptsize $\pm 0.36$}  &  $91.23$ {\scriptsize $\pm 0.24$}  &  $78.67$ {\scriptsize $\pm 0.24$}  \\
            3            &  $90.31$ {\scriptsize $\pm 0.46$}  &  $91.34$ {\scriptsize $\pm 0.22$}  &  $79.02$ {\scriptsize $\pm 0.18$}  \\
            4            &  $90.30$ {\scriptsize $\pm 0.44$}  &  $91.40$ {\scriptsize $\pm 0.22$}  &  $79.16$ {\scriptsize $\pm 0.21$}  \\
        \bottomrule
    \end{tabular}
\end{table}

As another aspect of the iterative mixing layers, we also perform scratch training of the Mixer models with a multiple number of iterations for all the mixing layers.
Here, we adopt the number of iterations for all the mixing layers from one to four, and keep the rest of the training configurations intact.
\tref{tab:iter-cifar} shows the results.
The results for a single iteration indicate the baseline from \tref{tab:result-cifar}.
We observe that the performance of \psmixer ~monotonically improves with a large number of iterations, while the performance improvement of \vmixer ~stops immediately.
In particular, there is a clear margin between \pmixer ~and \vmixer ~for a large number of iterations.

\textbf{Ablation study.}
We here consider some ablation studies to examine the functions of the proposed models.
\tref{tab:ablation} shows the overall results.
We did not include bias terms in the two-layer MLPs of token- and channel-mixing blocks in \psmixer ~due to the correspondence with the Hopfield network side, while \vmixer ~does have.
\tref{tab:ablation-bias} tells us that adding the bias terms does not affect performance much, as \pmixer ~with bias terms and \vmixer ~differ only in the parallel or serial mixing blocks, as depicted in Figs.\ref{fig:paramixer} and \ref{fig:vanillamixer}.
As in \tref{tab:result-cifar}, we have observed enormous performance drops in \smixer.
One might expect that the small number of parameters causes such a performance drop, since \smixer ~has only roughly half the number of parameters as \pmixer.
(The number of parameters for each model is provided in \appenref{appendix:training}.)
We train \pmixer ~with half widths (Para (hw)) and \smixer ~with double widths (Sym (dw)) to match the parameter number condition, and keep the remaining configurations unchanged.
\begin{table}[ht]
    \caption{
        Ablation studies for the Mixer models trained with CIFAR-10 from scratch.
        Each entry corresponds to the top-1 accuracy (\%) of the model.}
    \label{tab:ablation}
    \begin{minipage}[b]{0.49\linewidth}
        \centering
        \subcaption{add bias}
        \label{tab:ablation-bias}
        \begin{tabular}{lc}
            \toprule
                Model       &  Top-1 acc.                        \\
            \midrule
                Vanilla     &  $89.61$ {\scriptsize $\pm 0.22$}  \\
                Para + bias &  $89.77$ {\scriptsize $\pm 0.19$}  \\
                Para        &  $89.80$ {\scriptsize $\pm 0.29$}  \\
                Sym         &  $77.86$ {\scriptsize $\pm 0.53$}  \\
            \bottomrule
        \end{tabular}
    \end{minipage}
    \begin{minipage}[b]{0.49\linewidth}
        \centering
        \subcaption{double/half the widths}
        \label{tab:ablation-hratio}
        \begin{tabular}{lc}
            \toprule
                Model     &  Top-1 acc.                        \\
            \midrule
                Para      &  $89.80$ {\scriptsize $\pm 0.29$}  \\
                Para (hw) &  $87.14$ {\scriptsize $\pm 0.33$}  \\
                Sym  (dw) &  $79.11$ {\scriptsize $\pm 0.36$}  \\
                Sym       &  $77.86$ {\scriptsize $\pm 0.53$}  \\
            \bottomrule
        \end{tabular}
    \end{minipage}  \\
    \begin{minipage}[b]{0.49\linewidth}
        \centering
        \subcaption{symmetric vanilla}
        \label{tab:ablation-vanillasym}
        \begin{tabular}{lc}
            \toprule
                Model      &  Top-1 acc.                        \\
            \midrule
                Vanilla    &  $89.61$ {\scriptsize $\pm 0.22$}  \\
                VanillaSym &  $81.80$ {\scriptsize $\pm 0.20$}  \\
                Para       &  $89.80$ {\scriptsize $\pm 0.29$}  \\
                Sym        &  $77.86$ {\scriptsize $\pm 0.53$}  \\
            \bottomrule
        \end{tabular}
    \end{minipage}
    \begin{minipage}[b]{0.49\linewidth}
        \centering
        \subcaption{channel-only layernorm}
        \label{tab:ablation-channelln}
        \begin{tabular}{lc}
            \toprule
                Model   &  Top-1 acc.                        \\
            \midrule
                Vanilla &  $91.03$ {\scriptsize $\pm 0.35$}  \\
                Para    &  $90.73$ {\scriptsize $\pm 0.21$}  \\
                Sym     &  $80.91$ {\scriptsize $\pm 0.21$}  \\
            \bottomrule
        \end{tabular}
    \end{minipage}
\end{table}
\tref{tab:ablation-hratio} shows that there is still a large gap between Para (hw) and Sym (dw).
This result indicates that the symmetry condition imposed on the weight matrices of mixing layers contributes more crucially to the classification performance than the number of parameters.
In addition, this trend is observed not only between \psmixer ~but also in \vmixer ~(\tref{tab:ablation-vanillasym}).
This may imply that the symmetry breaking of weight matrices in the mixing modules of MetaFormers could generically be crucial to ensuring performance.
Finally, to examine the effect of the unusual layer normalization within the Mixer models, we replace the normalization layer with the ordinary channel-only layer normalization.
From \tref{tab:ablation-channelln}, we see that all the Mixer models achieve slightly better performance.
This is probably because our layer normalization normalizes the features over both token and channel axes, causing the sequence of token vectors to be overly normalized, thus suppressing individuality.

%% file: sec6.tex
\section{Symmetry Breaking}
\label{sec:asmixer}

From the discussion in the previous section, we observed that the symmetry condition imposed on the weight matrices of mixing layers is actually a constraint for the image recognition task.
In this section, we explore the effect of symmetry breaking of weight matrices in the context of Hopfield networks, and investigate its implications for Mixer models.

\subsection{A formulation}

Hopfield networks, as associative memory models, are generally supposed to have symmetric weight matrices between neuron layers, as discussed in the previous sections.
In order to examine the effect of symmetry breaking of weight matrices, we propose a minimal extension of the setup of the three-layer \kro.
In doing so, it is natural to extend the original energy function \eref{eq:energy-kro} to the following interaction-symmetric form,
\begin{align}
    &E(\xs, \xv, \xc)
        = \sum_{A=s,v,c} \left( \left( x^A \right)^\top g^A - L^A \right)  \nonumber  \\
    &\quad - \frac12 \left( \left( \gv \right)^\top \xii{v}{s} \gs + \left( \gs \right)^\top \xii{s}{v} \gv \right)
        - \frac12 \left( \left( \gc \right)^\top \xii{c}{v} \gv + \left( \gv \right)^\top \xii{v}{c} \gc \right),
\end{align}
where the arguments for $g^A$ and $L^A$ are again omitted.
If $\xii{v}{s}=\left( \xii{s}{v} \right)^\top$ and $\xii{v}{c}=\left( \xii{c}{v} \right)^\top$, this energy function is nothing but the original one.

We then extend the hidden-to-visible interactions to include a symmetry breaking term,
\begin{equation}
    \xii{v}{s} = \left( \xii{s}{v} \right)^\top + \axii{v}{s},
    \quad
    \xii{v}{c} = \left( \xii{c}{v} \right)^\top + \axii{v}{c}.
    \label{eq:asym-weights}
\end{equation}
Taking the corresponding Lagrangians for \psmixer, and considering the adiabatic limit, $\tv \gg \ts, \tc$, the total energy with symmetry breaking terms reads:
\begin{multline}
    E(\xv)
        = {\xv}^{\top} \LN(\xv)  -  \sqrt{ \sum_{i,I} (\xv_{iI} - \bar{v})^2 }  \\
        - \sum_{\alpha=1}^{\Ns} \igelu  \left( \xs_\alpha \right) + \frac12 \sum_{i,I,\alpha} \LN(\xv)_{iI} \axii{v}{s}_{i\alpha} \gelu(\xs_\alpha)  \\
        - \sum_{\beta=1}^{\Nc}  \igelu  \left( \xc_\beta \right)  + \frac12 \sum_{i,I,\beta}  \LN(\xv)_{iI} \axii{v}{c}_{I\beta}  \gelu(\xc_\beta).
    \label{eq:energy-asym}
\end{multline}
It is understood that $\xs$ and $\xc$ above are shorthands for \eref{eq:sc-red}.
We refer to this extended energy function as the pseudo energy function.
We defined the Lagrangians for the hidden neurons as $L^s=L^c=\sum_a \igelu(x^A_a)$, where $\igelu$ is the primitive function of $\gelu$:
\begin{equation}
    \igelu(z) = \frac14 \left( z^2 + (z^2 -1 )\operatorname{erf}\left( \frac{z}{\sqrt{2}} \right) + z\sqrt{\frac{2}{\pi}} e^{-\frac{z^2}{2}} \right),
    \quad
    \igelu' = \gelu.
\end{equation}

\subsection{Degeneracy resolution}

With these Lagrangians and the pseudo energy function, the time evolution of the visible neurons $\xv$ is governed by the differential equation,
\begin{align}
    \dd{\xv(t)}{t}  
    &= \xii{v}{s} \gelu\left( \xii{s}{v} \LN(\xv(t)) \right)  \nonumber  \\
    &+ \xii{v}{c} \gelu\left( \xii{c}{v} \LN(\xv(t)) \right)
    - \xv(t),
    \label{eq:asym-dynamical}
\end{align}
where we set $\tv =1$.
In this equation, the weight matrices from hidden neurons to visible neurons generically involve the symmetry breaking term as in \eref{eq:asym-weights}.
We conduct an empirical study of the effects of symmetry breaking on associative memories by solving this dynamical equation using the \texttt{torchdiffeq} framework \cite{chen2018neuralode,torchdiffeq}.

\begin{figure}[ht]
    \begin{minipage}[b]{0.33\linewidth}
        \centering
        \includegraphics[keepaspectratio, scale=0.24]{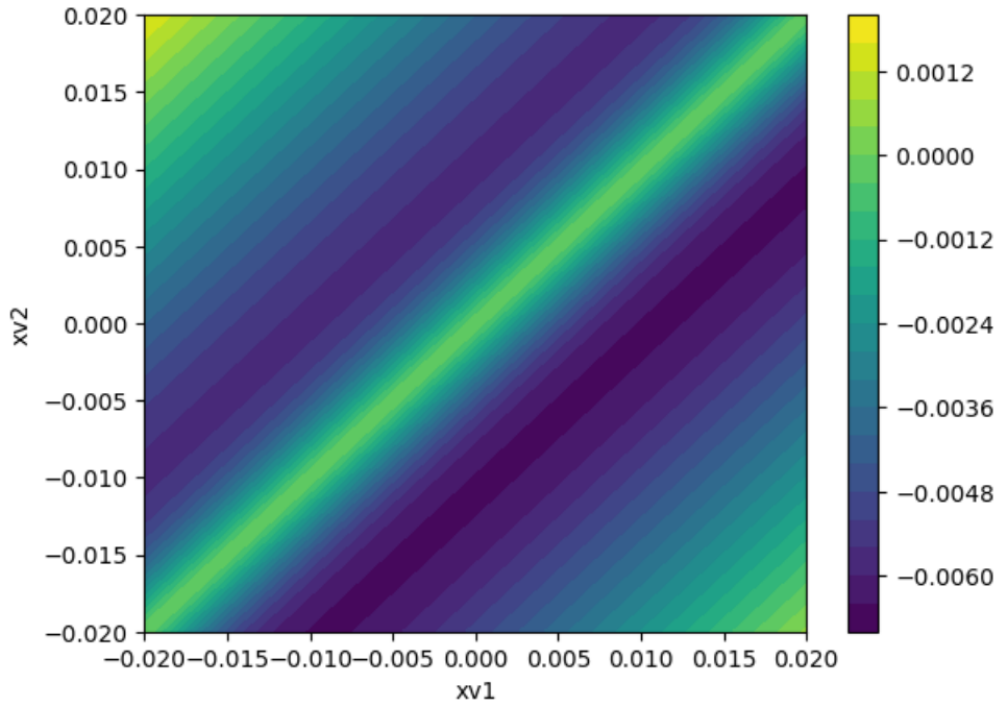}
        \subcaption{}
        \label{fig:energy-sym_2d}
    \end{minipage}
    \begin{minipage}[b]{0.33\linewidth}
        \centering
        \includegraphics[keepaspectratio, scale=0.32]{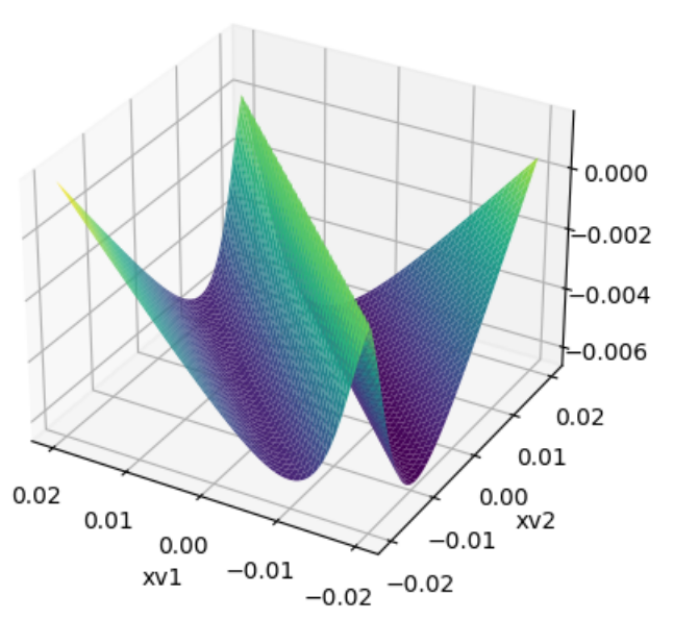}
        \subcaption{}
        \label{fig:energy-sym_3d}
    \end{minipage}
    \begin{minipage}[b]{0.33\linewidth}
        \centering
        \includegraphics[keepaspectratio, scale=0.24]{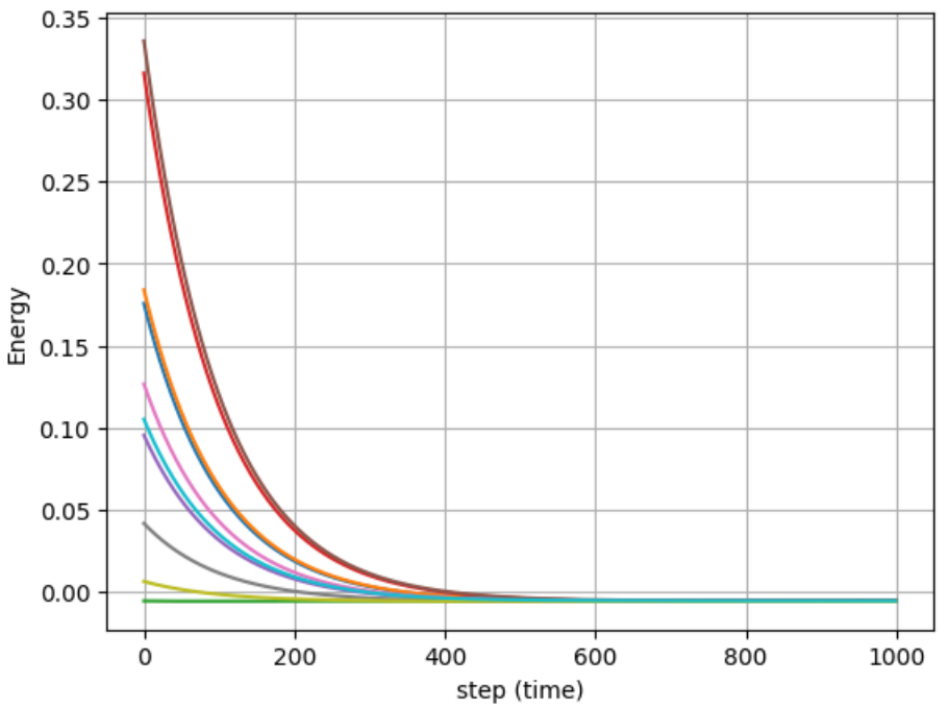}
        \subcaption{}
        \label{fig:energyt-sym_toy}
    \end{minipage}
    \caption{
        The pseudo energy with symmetric weights.
        (a), (b) Energy landscape for $N_v =2$.
        }
    \label{fig:landscape-sym}
\end{figure}

\begin{figure}[ht]
    \begin{minipage}[b]{0.33\linewidth}
        \centering
        \includegraphics[keepaspectratio, scale=0.24]{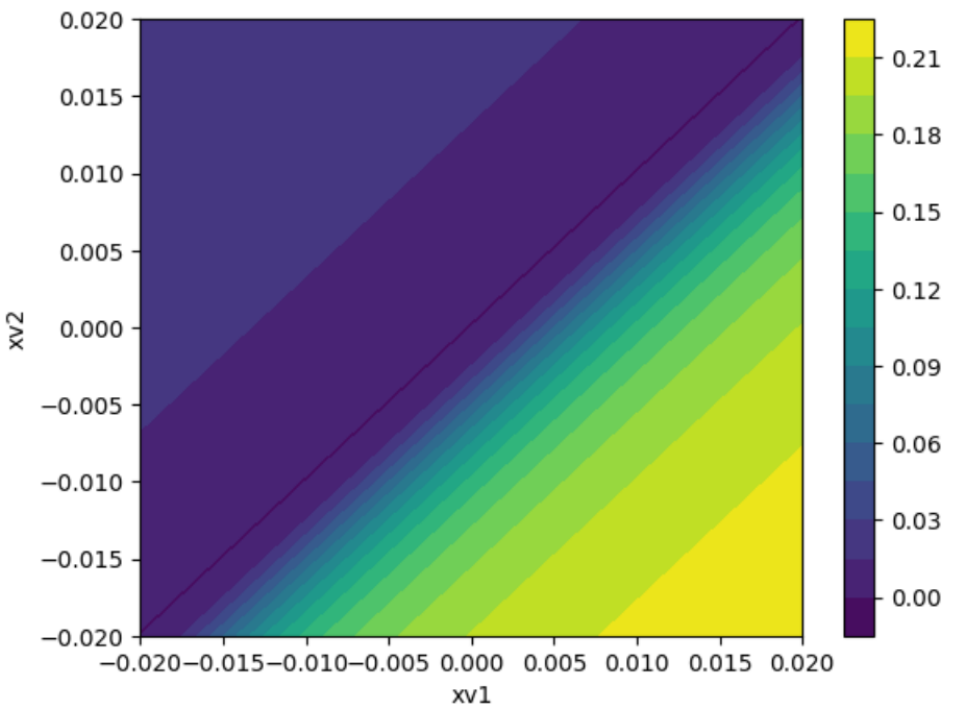}
        \subcaption{}
        \label{fig:energy-asym_2d}
    \end{minipage}
    \begin{minipage}[b]{0.33\linewidth}
        \centering
        \includegraphics[keepaspectratio, scale=0.32]{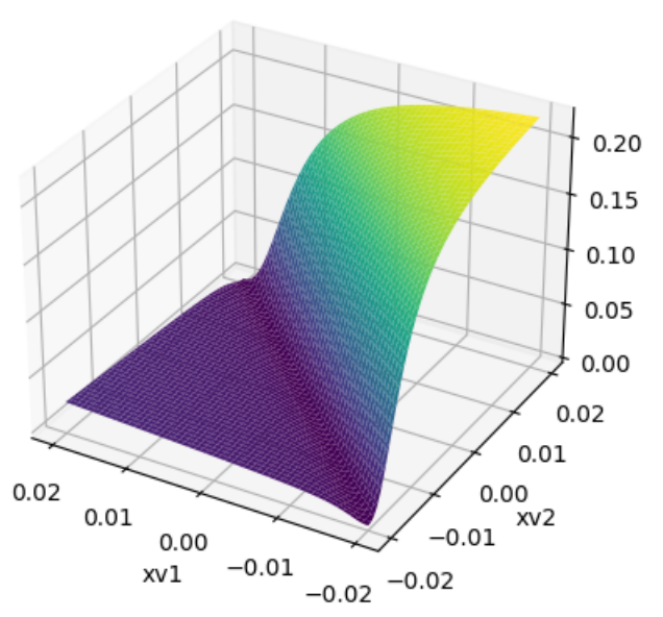}
        \subcaption{}
        \label{fig:energy-asym_3d}
    \end{minipage}
    \begin{minipage}[b]{0.33\linewidth}
        \centering
        \includegraphics[keepaspectratio, scale=0.24]{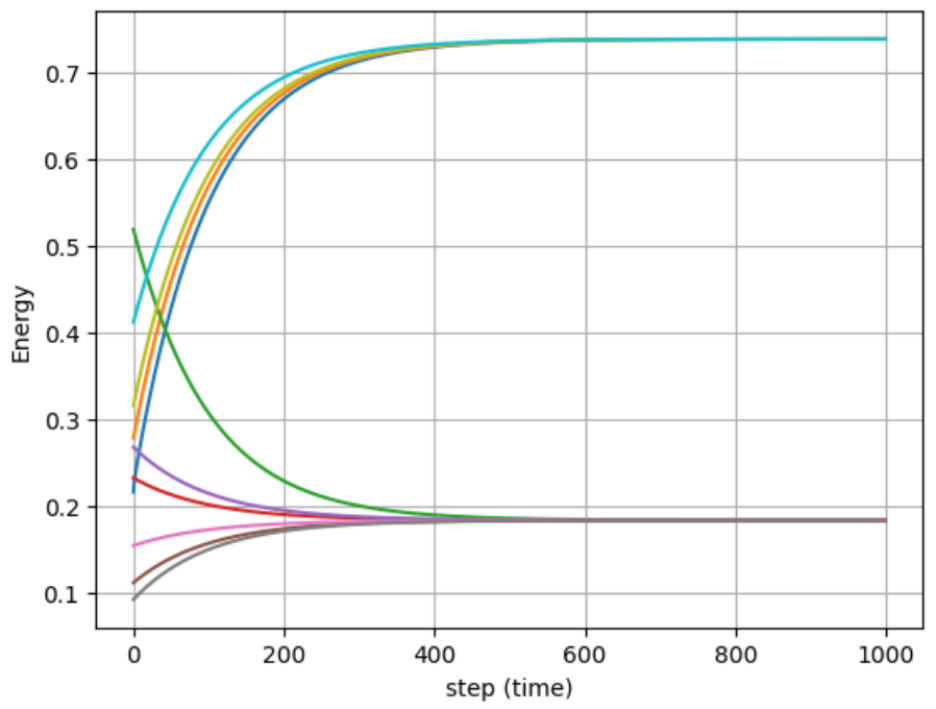}
        \subcaption{}
        \label{fig:energyt-asym_toy}
    \end{minipage}
    \caption{
        (a), (b) Pseudo energy landscape for $N_v =2$.
        }
    \label{fig:landscape-asym}
\end{figure}

Let us first take a simple example of $N_{v_s} = 2$ and $N_{v_c}=1$, for which the pseudo energy function \eref{eq:energy-asym} can be plotted in 2D/3D.
A random initialization of the weight matrices with the symmetric condition, $\axii{v}{\cdot}=0$, yields an energy landscape as shown in Figs.~\ref{fig:energy-sym_2d} and \ref{fig:energy-sym_3d}.
The energy function becomes symmetric with respect to $\xv_1 + \xv_2 = 0$, as in this simple case the Lagrangian for $\xv$ reduces to $L^v=\abs{\xv_1 - \xv_2}/\sqrt{2}$.
The minima of this energy function form two zero modes along the $\xv_1 = \xv_2$ line.
Any solution trajectories in this system approach the degenerate minima of the energy function (\fref{fig:energyt-sym_toy}).
By turning on the symmetry breaking terms $\axii{v}{\cdot}$, the degeneracy of the attractors is resolved and solution trajectories can retrieve different patterns in turn (\fref{fig:landscape-asym}).
It is noted that for the dynamical equation (\ref{eq:asym-dynamical}) with the symmetry breaking terms, the pseudo energy function does not necessarily decrease along the trajectories.
These observations imply that the associative memories the system stores degenerate for the symmetric weights, leading to an extremely limited effective memorization capability.

\begin{figure}[ht]
    \begin{minipage}[b]{0.495\linewidth}
        \centering
        \includegraphics[keepaspectratio, scale=0.32]{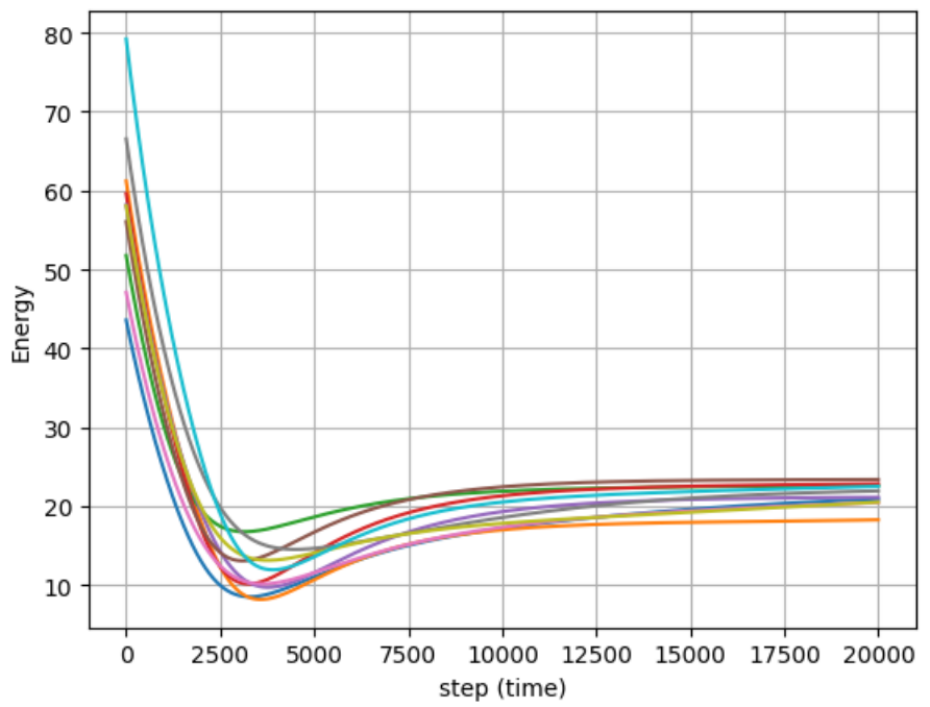}
        \subcaption{$\axii{v}{\cdot}=0$}
        \label{fig:energyt-sym}
    \end{minipage}
    \hspace{-1em}
    \begin{minipage}[b]{0.495\linewidth}
        \centering
        \includegraphics[keepaspectratio, scale=0.32]{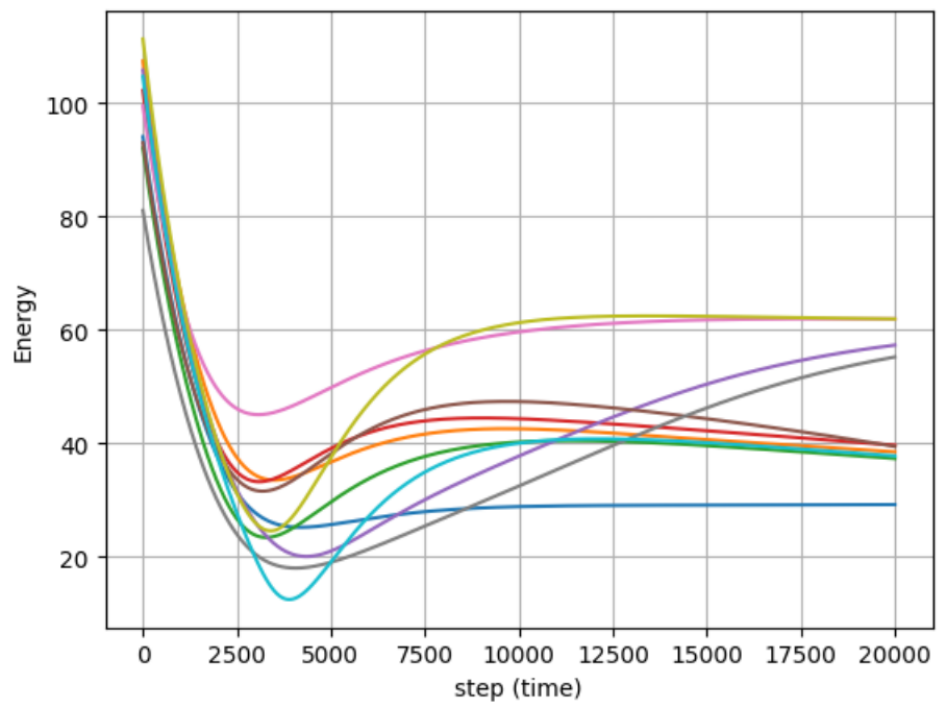}
        \subcaption{$\axii{v}{\cdot}\neq 0$}
        \label{fig:energyt-asym}
    \end{minipage}
    \caption{
        Time evolution of the pseudo energy along solution trajectories for the cases of (a) symmetric weight and (b) involving the symmetry breaking term.
        A slight increase of energy for (a) is due to $\igelu$.}
    \label{fig:energyt}
\end{figure}

We can gain more insights into this system by increasing the number of neurons.
For $N_{v_s}=4$, $N_{v_c}=8$, $\Ns=20$, and $\Nc=160$, we obtain \fref{fig:energyt}, although the full landscape of the pseudo energy function cannot be visualized.
We observe that $N_{v_s}$ roughly corresponds to the number of attractors, $N_{v_c}$ lifts the total energy, and a large total number of neurons makes the convergence slower.
These observations are natural, as $N_{v_s}$ corresponds to the number of tokens and $N_{v_c}$ to the number of channels in terms of Mixer models of deep neural networks.

\subsection{\asmixer}
\label{sec:asymmixer}

Having knowledge from the Hopfield network side, we consider the symmetry breaking of \smixer ~by stacking the discrete version of the previous subsection as mixing layers.
We refer to the parallelized MLP-Mixer with asymmetric weights as the \emph{\asmixer}, whose mixing layers consist of a symmetry-breaking extension of \eref{eq:symmixer},
\begin{align}
    W_2 = W_1^{\top} + \tilde{W}_2,  \quad  W_4 = W_3^{\top} + \tilde{W}_4.
\end{align}

To perturbatively add the symmetry breaking and study its effect on the trainability of the model, we train the \asmixer ~with the following custom loss:
\begin{equation}
    \Ls = \Ls_{\text{CE}} + \reg \sum_{l=1}^{L} \sum_{a=2,4} \norm{\tilde{W}_a^l}_F^2,
\end{equation}
where $\Ls_{\text{CE}}$ is the cross-entropy loss function, $L$ is the number of layers, and $\norm{\cdot}_F$ denotes the Frobenius norm.

\begin{figure}[t]
    \begin{minipage}[b]{0.495\linewidth}
        \centering
        \includegraphics[keepaspectratio, scale=0.3]{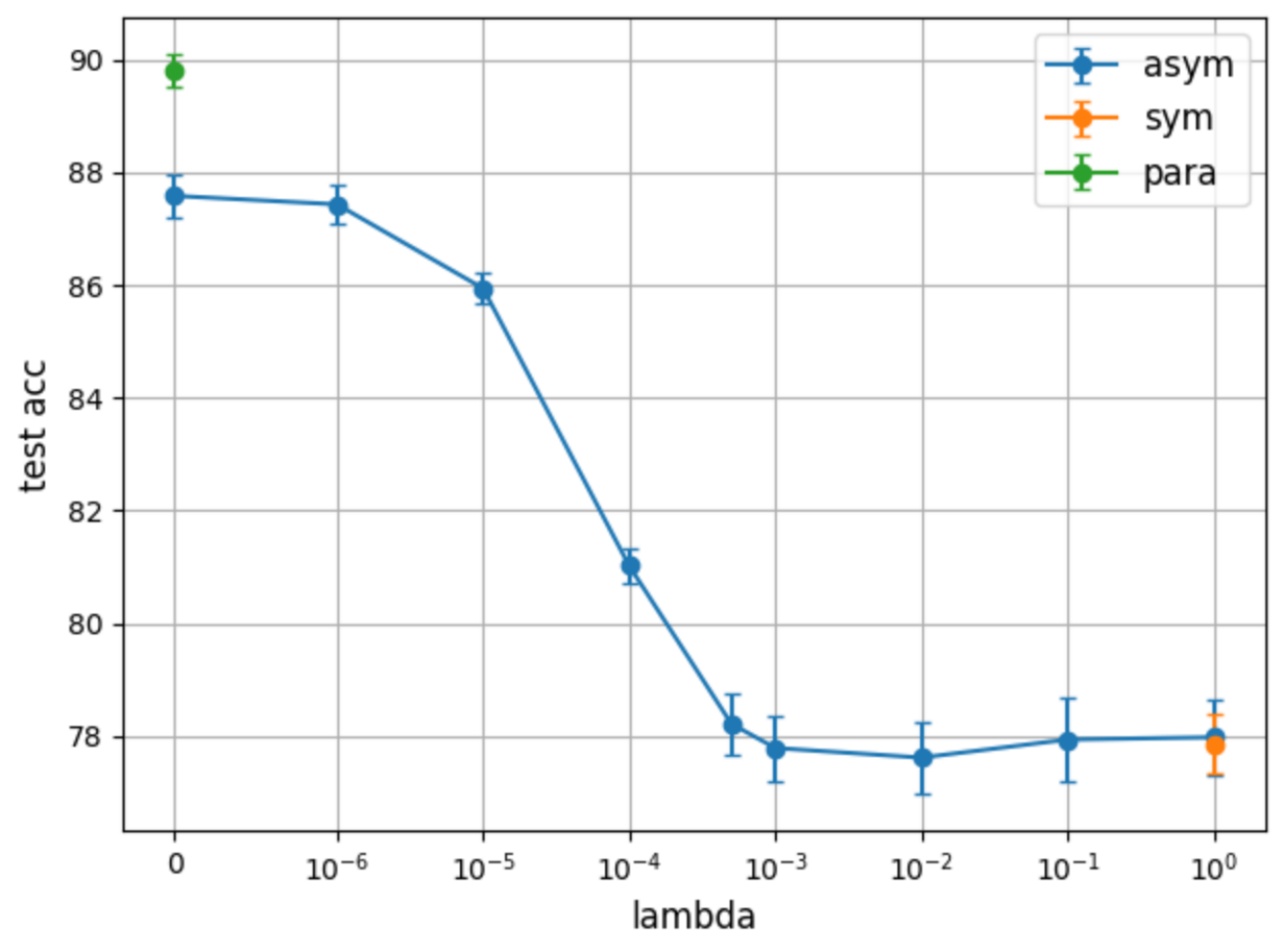}
        \subcaption{}
        \label{fig:acc-lambda}
    \end{minipage}
    \begin{minipage}[b]{0.495\linewidth}
        \centering
        \includegraphics[keepaspectratio, scale=0.3]{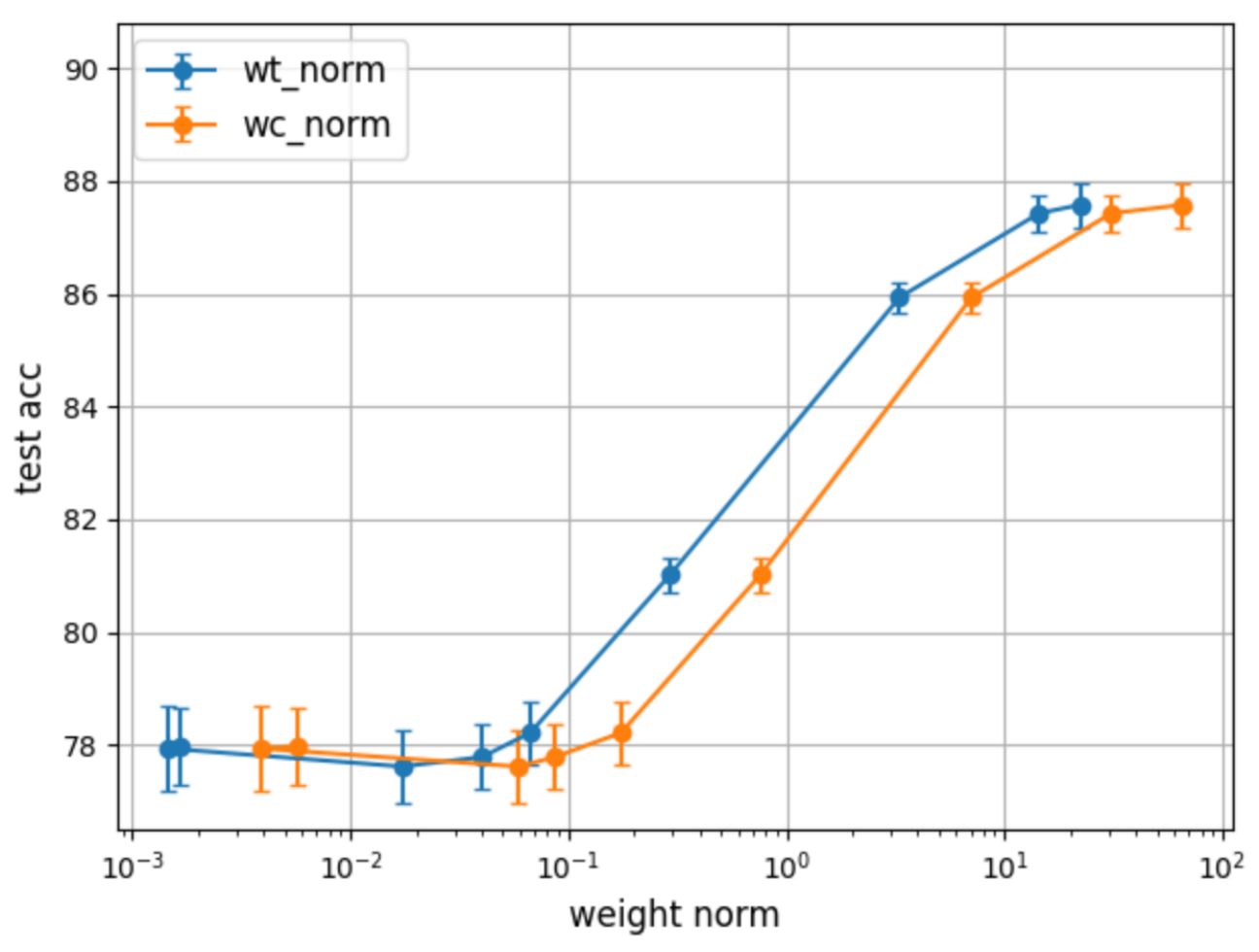}
        \subcaption{}
        \label{fig:acc-weight}
    \end{minipage}
    \caption{
        Top-1 accuracy of \asmixer s compared with \pmixer ~and \smixer, trained with CIFAR-10 from scratch.
        (a) Performance transition of \asmixer.
        (b) Performance transition with respect to the norm of weights.
            \texttt{wt\_norm} and \texttt{wc\_norm} are norms of $\tilde{W}_2^L$ and $\tilde{W}_4^L$, respectively.}
    \label{fig:results-asym}
\end{figure}

\Figref{fig:results-asym} shows the results.
We use CIFAR-10 to train the \asmixer s from scratch with $\reg = \set{0, 10^{-r}}$ for $r=0, \dots, 6$.
For $\reg=1$ ($r=0$), the symmetry breaking terms $\tilde{W}$ are almost not allowed to have non-trivial entries due to the penalty term in the loss function, for which the trained \asmixer ~boils down to \smixer.
For $\reg=0$, in contrast, $\tilde{W}$ have no restrictions to learn and thus the model weights are trained in a similar manner as \pmixer.
\Figref{fig:acc-lambda} provides reasonable results for the trainability of \asmixer s, interpolating the performance of \smixer ~and \pmixer.
The discrepancy between the performance of \asmixer ~with $\reg=0$ and \pmixer ~is due to the architecture designs; we assume that the symmetry breaking terms in \asmixer ~are included only in the hidden-to-visible interactions for simplicity.
These observations demonstrate consistent results with the previous subsection, indicating that symmetric weights in Mixer models have only a small amount of effective associative memories, and that symmetry breaking plays a crucial role in assuring performance.

%% file: sec7.tex
\section{Discussion}

We proposed a novel correspondence between Hopfield networks and MLP-Mixer models, and identified the \psmixer ~as a stack of associative memory models.
The proposed models consist of the parallelized mixing layer, which naturally includes the symmetric layer normalization module from the Hopfield network side.
We demonstrated the basic properties of the models as associative memories and examined their visual recognition capabilities in the context of deep neural networks.
From the numerical experiments, we observed that the symmetry conditions on interaction matrices in Hopfield networks are indeed a constraint on the performance of Mixer models.
The empirical studies imply that Mixer models with symmetric weights have a highly restricted effective memorization capacity and that symmetry breaking essentially plays a crucial role in ensuring performance.

One limitation of this paper is its narrow focus on the Lagrangians in the formulation of the proposed models.
The Lagrangians for hidden neurons (in other words, activation functions $\gs$ and $\gc$) are actually not restricted in the derivation of the \psmixer.
Reconsideration of the Lagrangians for hidden neurons might provide more insights into understanding the role of mixing modules in MetaFormers.
Another limitation is the lack of examination of applications to practical problems.
While the performance of \smixer ~significantly drops, as observed in \sref{sec:experiments}, there could be a task domain suitable for \smixer ~such as denoising or image retrieval, where Hopfield networks would inherently be effective.
In addition, it is worth pursuing the quantitative or exact analysis of the memorization capacity of \emixer ~(\sref{sec:emixer}) and its symmetry breaking phase (\asmixer, \sref{sec:asmixer}).
To do so, in a similar fashion to \cite{lucibello2024exponential,yampolskaya2023controlling}, it would be helpful to utilize tools developed in statistical physics.
We leave these aspects of the study for future exploration.

%% file: appendix.tex
\section{Some Useful Formulae}
\label{appendix:formulae}

The choice of Lagrangian $L^A: \R^{N_A}\to \R$ defines the activation function $g^A: \R^{N_A}\to \R^{N_A}$ such that $g^A = \nabla L^A$.
These functions introduce non-linearities into the dynamics of neuron layers.
We here show three examples used in the main text.
For other common examples, see e.g., \cite{hoover2022universal}.

\textbf{LayerNorm.}
For $x\in \R^N$, we obtain the layer normalization module from the following Lagrangian,
\begin{equation}
    L_{\text{LN}}(x) = D\gamma \sqrt{\frac1D \sum_{i=1}^{N} \left( x_i - \bar{x} \right)^2 + \epsilon} + \sum_i \delta_i x_i,
\end{equation}
where $\bar{x}=\sum_i x_i / N$, $\gamma$ and $\delta_i$ are learnable parameters, $\epsilon$ is a regularization constant, and $D$ is an arbitrary constant.
The derivative of this Lagrangian computes
\begin{align}
    g_i(x)
        = \dpdp{L_{\text{LN}}(x)}{x_i}
        &= \gamma \frac{x_i - \bar{x}}{\sqrt{\frac1D \sum_j \left( x_j - \bar{x} \right)^2 + \epsilon}} + \delta_i  \nonumber  \\
        &= \LN(x)_i.
\end{align}
One can see that by taking $\gamma=D=1$ and $\delta_i=\epsilon=0$, this Lagrangian and the activation function give those discussed in the main text.

\textbf{ReLU.}
For $x\in \R^N$, the Lagrangian
\begin{equation}
    L_{\text{ReLU}}(x) = \frac12 \sum_{i=1}^{N} \max(x_i,0)^2
\end{equation}
gives the ReLU activation function, which acts on $x$ element-wise:
\begin{equation}
    g_i(x) = \dpdp{L_{\text{ReLU}}(x)}{x_i} = \max(x_i,0).
\end{equation}

\textbf{GELU.}
For $x\in \R^N$, we obtain the GELU activation function from the Lagrangian,
\begin{align}
    L_{\text{GELU}}(x)
        &= \sum_{i=1}^{N} \igelu(x_i),  \\
    \igelu(z)
        &= \frac14 \left( z^2 + (z^2 -1 )\operatorname{erf}\left( \frac{z}{\sqrt{2}} \right) + z\sqrt{\frac{2}{\pi}} e^{-\frac{z^2}{2}} \right).
\end{align}
One finds that $\igelu$ is a primitive function of the GELU activation function, acting on $x$ element-wise,
\begin{equation}
    g_i(x) = \dpdp{L_{\text{GELU}}(x)}{x_i} = \frac{x_i}{2}\left( 1 + \operatorname{erf}\left( \frac{x_i}{\sqrt{2}} \right) \right).
\end{equation}

\section{Experimental Details}
\label{appendix:experiments}

\subsection{Code of parallelized mixing layer}

The pseudo-code of the parallelized mixing layer is shown in \algref{alg:pseudocode}.
As mentioned in the main text, we use PyTorch Image Models \texttt{timm} \cite{rw2019timm}%
\footnote{%
    \url{https://github.com/huggingface/pytorch-image-models}.
}
for the implementation of the models.
The \psmixer ~models have mostly the same structure as the ordinary MLP-Mixer and no additional hyperparameters.
The two main differences from the ordinary Mixer are the parallelized token- and channel-mixing modules and the symmetric layer normalization applied along both token and channel axes.

\begin{algorithm}
\caption{Pseudo-code of the parallelized mixing layer, PyTorch-like code.}
\label{alg:pseudocode}
\begin{lstlisting}[language=Python]
class pMixerBlock(nn.Module):
    def __init__(self, dim, seq_len, h_r=1, n_iter=1):
        super().__init__()
        d_t, d_c = [int(x * dim) for x in to_2tuple((0.5,4.0))]
        self.mlp_t = Mlp(seq_len, int(d_t*h_r), act_layer=nn.GELU)
        self.mlp_c = Mlp(dim, int(d_c*h_r), act_layer=nn.GELU)
        self.drop_path = DropPath(0.1)
        self.n_iter = n_iter

        # symmetric layer normalization
        self.norm = nn.LayerNorm((seq_len, dim))

    # parallelized MLPs from the Hopfield/Mixer correspondence
    def forward(self, x):
        for _ in range(self.n_iter):
            x = x
                + self.drop_path(self.mlp_t(
                    self.norm(x).transpose(1, 2)).transpose(1, 2))
                + self.drop_path(self.mlp_c(self.norm(x)))
        return x
\end{lstlisting}
\end{algorithm}

\subsection{Training details}
\label{appendix:training}

We here report the detailed training setups commonly used for the Mixer models in Secs.~\ref{sec:experiments} and \ref{sec:asymmixer}.
We basically follow the previous study \cite{TouvronICML2021}, and also employ \cite{hou2022vision}%
\footnote{%
    \url{https://github.com/houqb/VisionPermutator}.
}
for some considerations.
The number of trainable parameters of each model is shown in \tref{tab:params}.

\begin{table}[ht]
    \centering
    \caption{Hyperparameters commonly used for the Mixer models for fair comparison.}
    \label{tab:hyperparameters}
    \begin{tabular}{lc}
        \toprule
        Training configuration &  Value         \\
        \cmidrule(rl){1-2}
        \# layers              &  8             \\
        optimizer              &  AdamW         \\
        training epochs        &  300           \\
        batch size             &  384           \\
        base learning rate     &  $5 \times 10^{-4}$  \\
        weight decay           &  0.05          \\
        optimizer $\epsilon$   &  $10^{-8}$     \\
        optimizer momentum     &  $\beta_1=0.9$, $\beta_2=0.999$  \\
        learning rate schedule &  cosine decay  \\
        lower learning rate bound & $10^{-6}$   \\
        warmup epochs          &  20            \\
        warmup schedule        &  linear        \\
        warmup learning rate   &  $10^{-6}$     \\
        cooldown epochs        &  10            \\
        crop ratio             &  0.875         \\
        RandAugment            &  (9, 0.5)      \\
        mixup $\alpha$         &  0.8           \\
        cutmix $\alpha$        &  1.0           \\
        random erasing         &  0.25          \\
        label smoothing        &  0.1           \\
        stochastic depth       &  0.1           \\
        \bottomrule
    \end{tabular}
\end{table}

\begin{table}[ht]
    \centering
    \caption{Number of parameters of the Mixer models.}
    \label{tab:params}
    \begin{tabular}{lcccc}
        \toprule
                      & \vmixer & \pmixer & \smixer & \asmixer \\
        \cmidrule(rl){2-5}
            \# params & 21.2 M  & 19.6 M  & 10.8 M  & 19.6 M   \\
        \bottomrule
    \end{tabular}
\end{table}